\documentclass[10pt,twocolumn,letterpaper]{article}

\usepackage[pagenumbers]{cvpr} 

\definecolor{cvprblue}{rgb}{0.21,0.49,0.74}
\usepackage[pagebackref,breaklinks,colorlinks,allcolors=cvprblue]{hyperref}
\usepackage{algorithm,algorithmic}
\usepackage{booktabs}
\usepackage{multirow}
\usepackage{amssymb}
\usepackage{pifont}
\usepackage{multirow}
\usepackage{amssymb}
\usepackage{bbding}
\usepackage{pifont}
\usepackage{makecell}
\usepackage[table,xcdraw]{xcolor}
\usepackage[normalem]{ulem}
\usepackage{graphicx}
\usepackage{bbold}

\title{SurfSLAM: Sim-to-Real Underwater Stereo Reconstruction For Real-Time SLAM}

\author{
Onur Bagoren$^*$ \and
Seth Isaacson$^*$ \and
Sacchin Sundar \and
Yung-Ching Sun \and
Anja Sheppard \and \\
Haoyu Ma \and \\
Abrar Shariff \and \\
Ram Vasudevan \and \\
Katherine A. Skinner
}
\usepackage[dvipsnames]{xcolor}

\usepackage{xspace}

\renewcommand{\L}{\mathcal{L}}
\newcommand{\R}{\mathbb{R}}
\renewcommand{\b}[1]{\mathbf{#1}}
\renewcommand{\cal}[1]{\mathcal{#1}}

\newcommand{\algname}{SurfSLAM\xspace}

\newif\ifreviewson
\newif\ifcommentson
\commentsonfalse
\reviewsonfalse

\newcommand{\Figure}{Figure~}

\newcommand{\Table}{Table~}

\newcommand{\SO}{\mathrm{SO}}
\newcommand{\SE}{\mathrm{SE}}
\newcommand{\SOthree}{\SO(3)}
\newcommand{\SEthree}{\SE(3)}

\newcommand{\World}{\mathtt{W}}
\newcommand{\NED}{\mathtt{N}}
\newcommand{\Imu}{\mathtt{I}}
\newcommand{\DVL}{\mathtt{D}}
\newcommand{\Barometer}{\mathtt{P}}
\newcommand{\Base}{\mathtt{{B}}}
\newcommand{\Camera}{\mathtt{C}}

\DeclareMathOperator*{\argmax}{argmax}
\DeclareMathOperator*{\argmin}{argmin}

\usepackage[font=small,labelfont=bf]{caption}
\begin{document}
\twocolumn[{
    \renewcommand\twocolumn[1][]{#1}
    \maketitle
    \vspace{-1.5em}
    \begin{center}
    \href{https://umfieldrobotics.github.io/SurfSLAM/}{\texttt{https://umfieldrobotics.github.io/SurfSLAM/}}
    \end{center}
    \vspace{0.5em}
    \centering
\includegraphics[width=\linewidth]{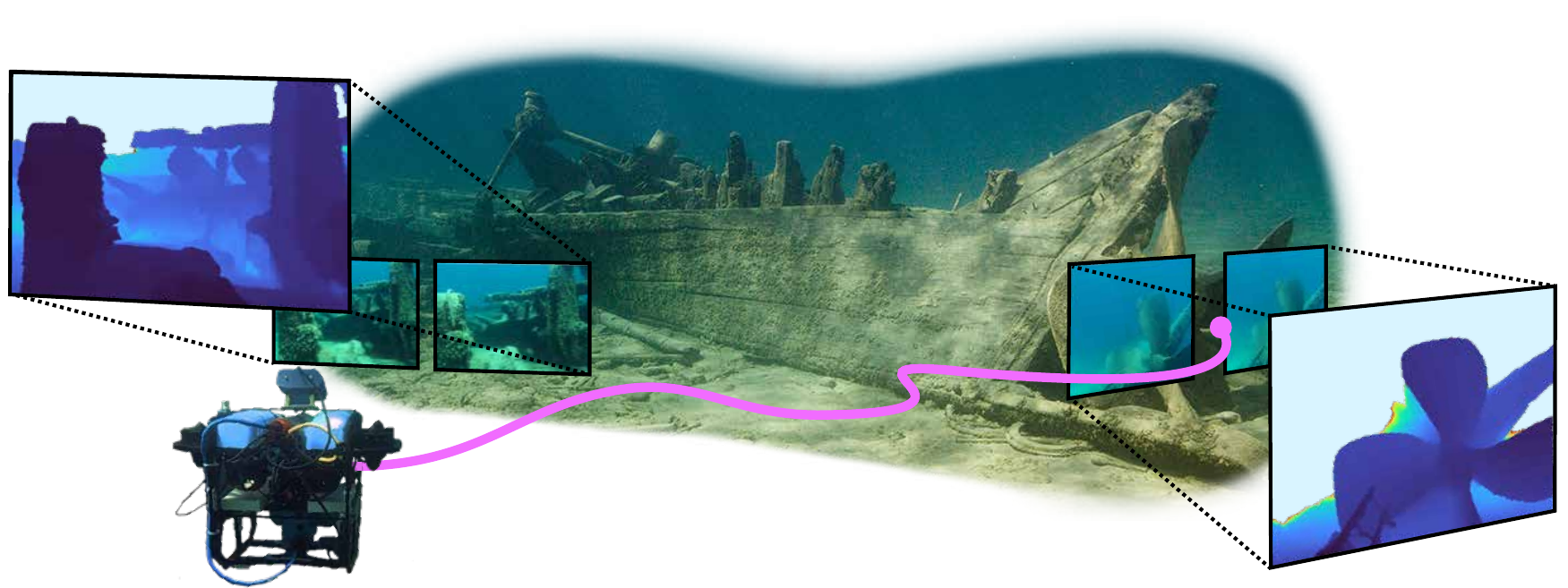}
    \captionof{figure}{\algname performs real-time SLAM in challenging underwater environments by fusing proprioceptive data with global registration. Accurate global registration is enabled by an underwater stereo disparity estimation algorithm that is fine-tuned via a novel sim-to-real pipeline. Illustrated is our BlueROV data collection platform navigating a shipwreck. Two stereo frames along the robot's path (drawn in pink) are visualized. Background image courtesy of the National Oceanic and Atmospheric Administration Thunder Bay National Marine Sanctuary.}
    \label{fig:fig1}
    \vspace{1em}
}]

\begingroup
\renewcommand\thefootnote{}%
\footnote{$^*$Denotes equal contribution}%
\footnote{This work was supported by the National Science Foundation under Award No. 2337774.}%
\footnote{All authors are with the Department of Robotics, University of Michigan, Ann Arbor, MI 48109, USA}%
\footnote{Corresponding author e-mail: {\tt\small obagoren@umich.edu}}
\addtocounter{footnote}{-1}%
\endgroup

\begin{abstract}
Localization and mapping are core perceptual capabilities for underwater robots.
Stereo cameras provide a low-cost means of directly estimating metric depth to support these tasks.
However, despite recent advances in stereo depth estimation on land, computing depth from image pairs in underwater scenes remains challenging.
In underwater environments, images are degraded by light attenuation, visual artifacts, and dynamic lighting conditions.
Furthermore, real-world underwater scenes frequently lack rich texture useful for stereo depth estimation and 3D reconstruction.
As a result, stereo estimation networks trained on in-air data cannot transfer directly to the underwater domain.
In addition, there is a lack of real-world underwater stereo datasets for supervised training of neural networks.
Poor underwater depth estimation is compounded in stereo-based Simultaneous Localization and Mapping (SLAM) algorithms, making it a fundamental challenge for underwater robot perception.
To address these challenges, we propose a novel framework that enables sim-to-real training of underwater stereo disparity estimation networks using simulated data and self-supervised finetuning.
We leverage our learned depth predictions to develop \algname, a novel framework for real-time underwater SLAM that fuses stereo cameras with IMU, barometric, and Doppler Velocity Log (DVL) measurements. 
Lastly, we collect a challenging real-world dataset of shipwreck surveys using an underwater robot. 
Our dataset features over 24,000 stereo pairs, along with high-quality, dense photogrammetry models and reference trajectories for evaluation. 
Through extensive experiments, we demonstrate the advantages of the proposed training approach on real-world data for improving stereo estimation in the underwater domain and for enabling accurate trajectory estimation and 3D reconstruction of complex shipwreck sites.
Code, real-world underwater data, and simulated training data will be made publicly available.
\end{abstract}

\section{Introduction}
Underwater simultaneous localization and mapping (SLAM) is a key challenge in marine robotics, with wide-ranging applications such as underwater inspection~\cite{hou2022underwater} and semantic mapping~\cite{singh2024opti}.
Vision plays a crucial role in this context.
Cameras are relatively low-cost sensors, and stereo cameras can provide metric depth information useful for reconstructing the scene geometry.
However, underwater imaging presents several fundamental differences from in-air imagery.
The presence of haze, attenuation, dynamic lighting, and suspended particles reduces contrast and detail.
Further, large areas of real-world underwater scenes often lack distinctive, trackable textures useful for estimating geometry~\cite{akkaynak2018arevised, Li_2017}.
These underwater effects can vary significantly between locations and conditions, making it challenging to develop perception systems that remain robust across diverse underwater environments~\cite{jerlov1968optical}.

\label{sec:intro}

Classical, hand-designed methods for stereo depth estimation struggle in textureless and hazy regions; as a result, they do not generalize well to the underwater domain~\cite{hirschmuller2007stereo}. 
For the in-air domain, significant attention has been devoted to the monocular depth estimation problem~\cite{yang2024depth}. In the underwater domain, several works have adapted in-air monocular depth estimation networks to account for underwater image characteristics ~\cite{yu2022udepth, ye2025uw, zhang2025spade}. 
Still, monocular depth estimation cannot reliably provide metric-scale depth.
Recent deep stereo networks achieve excellent performance in estimating in-air metric depth~\cite {wen2025foundationstereo, xu2025igev++, defom_stereo}.
However, as we demonstrate in this work, their accuracy degrades significantly underwater due to the domain shift induced by haze, attenuation, and dynamic lighting. 
Several works have proposed learning-based solutions to underwater depth estimation~\cite{ye2023underwater, skinner2019uwstereonet, wu2025stereoadapteradaptingstereodepth}, but a lack of foundation-scale underwater training data and limitations in simulation fidelity have limited their accuracy.

In this work, we propose \textbf{\algname}, illustrated in \Figure\ref{fig:fig1}. \algname is a novel framework for \textbf{S}im-to-real \textbf{U}nderwater stereo \textbf{R}econstruction \textbf{F}or real-time \textbf{SLAM}. 
Our sim-to-real stereo depth estimation pipeline leverages knowledge of underwater imaging effects to recover dense, reliable metric depth from underwater stereo imagery.
We demonstrate that our sim-to-real training enables state-of-the-art stereo depth estimation in the underwater domain.
As a result, the dense depth maps estimated by our approach provide strong geometric priors to improve navigation and mapping capabilities.
Our proposed SLAM framework fuses depth predictions with measurements from an IMU, Doppler Velocity Log (DVL), and barometer to track vehicle pose. 
The vehicle pose and dense depth maps are then used to estimate a dense 3D map of the scene.

Our method is evaluated on a dataset of long-duration shipwreck surveys for visual-acoustic-inertial SLAM. 
Our dataset includes reference 3D reconstructions and vehicle trajectories to support qualitative and quantitative evaluation. 
We present extensive experiments demonstrating that \algname improves both trajectory accuracy and reconstruction quality for underwater SLAM in challenging real-world environments compared to existing state-of-the-art methods. 
Code, model weights, real-world underwater data, and simulated training data will be made publicly available upon publication.

\section{Related Work} \label{sec:related}

\algname is a method that combines underwater SLAM, stereo depth estimation, and the transfer of deep learning methods to the underwater domain. We review the relevant literature below.

\subsection{Visual SLAM}
The task of estimating camera motion from stereo images, optionally fused with inertial sensing, has received enormous attention.
ORB-SLAM and its variants \cite{orbslam3} operate by matching and triangulating image features, tracking camera motion relative to feature maps with inertial priors, and jointly optimizing trajectories with a map via bundle adjustment \cite{orbslam3}.
These methods also implement a keypoint-based visual place recognition method that helps mitigate long-term drift.
While ORB-SLAM3 provides outstanding trajectory accuracy in visually rich scenes, it is prone to tracking loss in featureless regions and does not output a dense map useful for scene understanding.
As an alternative to feature-based methods, Droid-SLAM uses a learned optical flow module on images for dense correspondence matching and a learned pose update operator for incrementally estimating the state and producing a dense map representation~\cite{teed2021droid}.
Still, Droid-SLAM struggles in feature-sparse regions and lacks loop closure detection.
More recent approaches have applied large vision foundation models as the backbone to visual SLAM.
Prominent work leverages vision transformers (ViTs) pretrained on large-scale simulated in-air data for zero-shot feature matching or 3D reconstruction~\cite{leroy2024grounding,wang2025vggt}, with MASt3R-SLAM integrating MASt3R pointmaps into a monocular keyframe-based estimator~\cite{murai2025mast3r}.
VGGT-SLAM instead jointly estimates structure and motion on the $\mathtt{SL}(4)$ group from uncalibrated sequences~\cite{maggio2025vggt}.
These systems remain image-centric, heavily rely on ViT, and perform poorly in underwater domains where images fall out of the training distribution.
In contrast, our proposed method, \algname, fuses acoustic-inertial data with learned stereo depth estimation and visual place recognition to produce accurate trajectories in both featureless and feature-rich environments.

\subsection{Underwater Visual SLAM}
Early underwater visual SLAM focused on monocular-based navigation for large-scale inspections using pose-graph formulations~\cite{eustice_visually_2005, kim_combined_2011, kim_real-time_2013, shadow_contours}.
These works introduced informative visual constraint selection based on local geometry and demonstrated robust state estimation by fusing vision with DVL and inertial measurements for large-scale inspections~\cite{ozog_real-time_2013, ayoung_kim_pose-graph_2009}.
Later works studied how visual characteristics in underwater settings, such as directional light, could be leveraged within the visual front end, improving both tracking and mapping performance~\cite{shadow_contours}.

Later works tightly coupled acoustic and inertial sensors with visual information for more robust state estimation and mapping performance~\cite{xu_underwater_2021, vargas_robust_2021, avanthey_dense_2024}.
Pan et al. present a tightly integrated imaging sonar-IMU-camera system~\cite{pan_russo_2025}, while SVIn2 fuses a barometer and a mechanically scanning sonar into a sonar-visual-acoustic state estimator~\cite{svin2}.
Thoms et al. integrate a camera, DVL, IMU, and LiDAR for an autonomous surface vehicle, providing foundational work on the integration of the DVL into a pose-graph-based formulation~\cite{10149804}.
AQUA-SLAM similarly integrates a DVL into a visual-inertial odometry (VIO) framework for underwater applications, with online extrinsic calibration~\cite{xu2025aquaslamtightlycoupledunderwateracousticvisualinertial}.
These approaches yield robust pose estimates but produce sparse maps that lack the dense information available from stereo cameras present in each of the methods.

Song et al. produce dense maps through depth fusion of stereo depth maps along trajectories estimated by an acoustic-inertial tracker~\cite{song2024turtlmap}.
Wang et al. perform a similar process by using a Sum of Absolute Differences to compute stereo disparities, combined with poses estimated from SVIn2~\cite{wang_real-time_2023}.

Our work incorporates metric depth measurements estimated by a stereo network to form global registration factors, and combines them into an acoustic-visuo-inertial pose graph. As a result, \algname both maintains accurate tracking in featureless regions via acoustic-inertial fusion and mitigates long-term drift via global place recognition. 
Furthermore, with dense stereo depth estimation, SurfSLAM is able to output a dense map of complex underwater environments.

\subsection{Stereo Depth Estimation}\label{sss:dl_for_stereo}
Classical stereo and 3D reconstruction methods rely on hand-crafted pipelines for matching cost computation, disparity optimization, and refinement~\cite{scharstein2002taxonomy, sgbm, randall2023flsea, berman2020underwater}.
Commonly used methods include EpicFlow-based matching~\cite{berman2020underwater},~\cite{wang2015research}, and Agisoft Metashape~\cite{agisoft}.
These approaches tend to be highly sensitive to hyperparameters, which may need to be re-tuned depending on the scene.
Other methods focus on matching keypoints between images, then triangulating and optimizing 3D map points\cite{schoenberger2016sfm,schoenberger2016mvs}.
While this may provide good geometric accuracy, these methods cannot operate in textureless regions. Furthermore, they can take several hours to several days to run on real-world scenes, limiting their utility in the field.

\begin{figure*}[t!]
    \centering
    \includegraphics[width=1.0\linewidth]{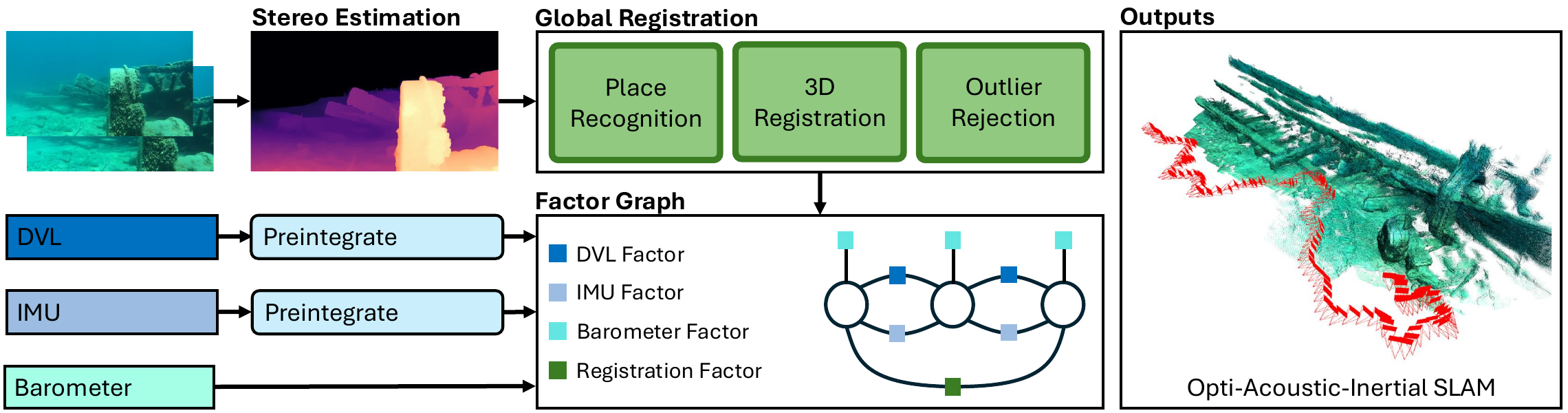}
    \caption{An overview of SurfSLAM. We take as input measurements from a barometer, IMU, DVL, and a stereo image pair. We maintain an acoustic-inertial pose graph that preintegrates measurements to maintain a pose throughout operation. In parallel, we use our finetuned underwater stereo network to produce metric depth maps. These depth estimates and stereo images are used to perform geometric tracking to perform global registration, and reduce drift over operation, producing accurate trajectories and dense maps during operation.}
    \label{fig:method_overview}
\end{figure*}
\newcommand{\x}{\mathbf{x}}
\newcommand{\disp}{\hat{\mathbf{d}}}

The introduction of deep learning for stereo depth estimation offers an alternative to hand-crafted approaches.
RAFT-Stereo~\cite{lipson2021raft} employs recurrent feature matching and iterative refinement with a typical encoder-decoder network architecture to produce disparity maps from a stereo image pair.
However, this approach often requires finetuning on a target domain for optimal performance.
The recent release of FoundationStereo~\cite{wen2025foundationstereo} has demonstrated the potential of models trained on large-scale simulated datasets to achieve excellent cross-domain, generalizable zero-shot performance on in-air data.
Foundation stereo leverages a pre-trained and frozen foundation model encoder from DepthAnything V2 \cite{depth_anything_v2}.
The foundation model features are fused with side-tuned CNN features and processed by a combination of convolutions, transformer mechanisms, and iterative refinement to produce high-quality disparity estimates \cite{wen2025foundationstereo}.
DEFOM-Stereo \cite{defom_stereo} also uses a foundation model to generate monocular depth estimation cues, but leverages a RAFT-Stereo backbone.
IGEV++ \cite{xu2025igev++} forgoes foundation model encoders and instead introduces multi-range geometry encoding volumes to capture different levels of encoding granularity, leading to strong performance in large-disparity scenes.
While IGEV++ cannot achieve the level of detail and accuracy of \cite{wen2025foundationstereo, defom_stereo}, it produces competitive results with significantly lower memory footprint and inference time than foundation model-based methods.
A prominent feature of deep learning methods for stereo prediction networks is the iterative refinement process, typically modeled as a variant of a gated recurrent unit (GRU) layer~\cite{wen2025foundationstereo} or an incremental disparity update step~\cite{defom_stereo}.
The iterative refinement step updates an initial disparity estimate based on correlation features extracted early in the network architecture, providing context relevant for refining disparity estimates.
Still, since these methods are developed and trained for in-air stereo depth estimation, they lack generalizability to underwater conditions such as turbidity, low light, caustics, and attenuation. 
The lack of large-scale training sets for training underwater stereo networks limits the ability to re-train these networks on real stereo imagery from underwater scenes.

\subsection{Deep Learning for Underwater Stereo}
A class of work that aims to overcome data scarcity in underwater stereo matching trains a network with direct supervision, using augmentations to make the training data more closely resemble the appearance of underwater images \cite{novel_stereo_style_transfer, lv2025uwstereo, ye2023underwater}.
In \cite{novel_stereo_style_transfer}, the authors employ a style transfer network to transform a reference underwater image into a target in-air stereo pair image with ground-truth disparity, and then train the network using these augmented images.
Other works have leveraged a physics-based approach, following underwater image formation models to augment large in-air datasets with water column effects~\cite{lv2025uwstereo}.

While simulated data mitigate the data scarcity problem, accurately simulating the complex underwater image formation remains challenging.
Self-supervised approaches aim to overcome this by leveraging stereo image pair consistency objectives.
UWStereoNet \cite{skinner2019uwstereonet} pre-trains an unsupervised Siamese network for disparity estimation on in-air data and then fine-tunes it on underwater data using a self-supervised objective of warping the stereo pairs onto each other and computing a photometric error between them.
More recently, StereoAdapter~\cite{wu2025stereoadapteradaptingstereodepth} targets underwater domain shift by parameter-efficiently adapting a monocular foundation encoder and a self-supervised objective that similarly minimizes a photometric error between stereo pairs with additional losses to handle occlusions and improve depths at edges.
In~\cite{wu2025stereoadapteradaptingstereodepth}, a dataset of 40,000 simulated images is also released, consisting of augmentations, texture maps and environments to better match the underwater image conditions, and perform pre-training with this dataset.

We propose a fusion of the two approaches. 
We first develop a novel, comprehensive train-time augmentation pipeline to generate realistic underwater images from large-scale in-air datasets, producing simulated data to train a model with direct supervision on the stereo disparity estimates.
This train-time augmentation is made possible due to the simulated data being coupled with the stereo intrinsic and extrinsic parameters, a key addition coupled with our simulated data.
As a fine-tuning approach, similar to \cite{skinner2019uwstereonet, wu2025stereoadapteradaptingstereodepth}, we employ a warping loss for self-supervised training. 
As a novel contribution over existing methods, we propose additional regularization to guide the model to make correct predictions when the background (i.e., water column) is present in the stereo image pairs.
The depths from the finetuned model are then used in our \algname framework to enable tracking and mapping in environments where the geometric cues allow for place recognition, global registration, and loop closures.
\section{Method Overview}
We propose \textbf{\algname}, a novel framework for \textbf{S}im-to-real \textbf{U}nderwater stereo \textbf{R}econstruction \textbf{F}or real-time \textbf{SLAM}. 
An overview of the proposed method is shown in \Figure\ref{fig:method_overview}. First, our sim-to-real stereo depth estimation pipeline leverages knowledge of underwater imaging effects to recover dense, reliable metric depth from underwater stereo imagery. 
Then, \algname fuses our dense stereo depth predictions using global registration with measurements from an IMU, DVL, and barometer for tracking vehicle pose. The poses and depth maps are then used to build a dense 3D reconstruction of the scene.

The discussion of the proposed technical approach is organized as follows: Subsection \ref{subsec:prelim} provides the notation and coordinate conventions used throughout the paper.
Section \ref{sec:stereo_depth} presents our proposed methodology for sim-to-real training of underwater stereo disparity estimation.
Then, Section \ref{sec:slam} describes our \algname system, which fuses stereo depth estimation with acoustic and inertial sensing for real-time localization and dense mapping of complex underwater scenes.

\subsection{Preliminaries}\label{subsec:prelim}

\subsubsection{Notation}
We denote scalars as lowercase italics ($x, y, z, \dots$), matrices as uppercase bold $\left(\b{X},\b{Y},\b{Z}, \dots\right)$ and vectors as lowercase bold $\left(\b{x}, \b{y}, \b{z}, \dots\right)$. 
Sensor measurements are denoted as capitalized calligraphic $\left(\cal{X}, \cal{Y}, \cal{Z}, \dots\right)$, 
and estimated variables are denoted with a bar $\bar{x}$.

\subsubsection{Frame Definitions}
The reference frames are shown in \Figure \ref{fig:frames}, where red indicates the $ x$-axis, green the $ y$-axis, and blue the $ z$-axis. The reference frames include the fixed world frame, $\World$, and the North-East-Down (NED) frame, $\NED$.
The base frame, $\Base$, and IMU frame, $\Imu$, are shown to be coincident, similar to that of ~\cite{Forster_2017}, and the sensor frames are represented as the DVL frame, $\DVL$, the barometer frame, $\Barometer$, and the left camera frame, $\Camera$.
All transformations and their notations follow the definitions in~\cite{transforms}.
\begingroup
\renewcommand{\arraystretch}{1.15}
\setlength{\extrarowheight}{1pt}
\newcommand{\yes}{\checkmark}
\newcommand{\no}{\ding{53}}
\newcommand{\pg}{PGM}
\newcommand{\at}{AprilTag}
\newcommand{\enc}{Encoder}
\renewcommand{\sim}{Sim.}
\newcommand{\two}[1]{\multirow{2}{*}{#1}}
\newcommand{\three}[1]{\multirow{3}{*}{#1}}
\newcommand{\four}[1]{\multirow{4}{*}{#1}}

\newcommand{\uwStereonet}{UWStereoNet~\cite{skinner2019uwstereonet}}
\newcommand{\svin}{SVIn2~\cite{svin2}}
\newcommand{\squid}{SQUID~\cite{berman2020underwater}}
\newcommand{\tank}{Tank~\cite{xu2025tank}}
\newcommand{\stereoAdapter}{StereoAdapter~\cite{wu2025stereoadapteradaptingstereodepth}}
\newcommand{\varos}{VAROS~\cite{zwilgmeyer2021varos}}
\newcommand{\uwslam}{UWslam~\cite{lizard}}
\newcommand{\flsea}{FLSea~\cite{randall2023flsea}}
\newcommand{\sotrue}{SOTRUE~\cite{sotrue25}}
\newcommand{\uwStereo}{UWStereo~\cite{lv2025uwstereo}}
\newcommand{\lidar}{\two{LiDAR}}

\begin{table*}
\centering
\footnotesize
\caption{A summary of underwater SLAM and stereo datasets. We propose two new datasets: a large simulated dataset with comprehensive underwater augmentation and a real-world dataset. \yes* denotes that the dataset will be released upon publication. Photogrammetry is shortened to PGM and simulation is shortened to Sim. for the ground truth column.}
\label{tab:datasets}
\begin{tabular}{lccccc|c|ccc|ccc}
\hline
                                                               &                                                                &                                                  &                               &                                                                        &                                                                             &                                & \multicolumn{3}{c|}{Water Effects}                                                         & \multicolumn{3}{c}{Additional Sensors}                                                     \\
                                                               & \multirow{-2}{*}{Name}                                         & \multirow{-2}{*}{Camera}                         & \multirow{-2}{*}{Splits}      & \multirow{-2}{*}{\begin{tabular}[c]{@{}c@{}}No.\\ Images\end{tabular}} & \multirow{-2}{*}{\begin{tabular}[c]{@{}c@{}}Publicly\\ Avail.\end{tabular}} & \multirow{-2}{*}{Ground Truth} & Turbid                       & Low  Light                   & Dir. Light                   & DVL                          & IMU                          & Baro.                        \\ \hline
\multicolumn{1}{l|}{}                                          &                                                                &                                                  & Train                         & 4,047                                                                  & \yes                                                                        &                                &                              &                              &                              &                              &                              &                              \\
\multicolumn{1}{l|}{}                                          & \multirow{-2}{*}{\uwStereonet}                                 & \multirow{-2}{*}{Stereo}                         & Test                          & 15                                                                     & \yes                                                                        & \multirow{-2}{*}{LiDAR}        & \multirow{-2}{*}{\yes}       &                              &                              &                              &                              &                              \\
\multicolumn{1}{l|}{}                                          & \squid                                                         & Stereo                                           & -                             & 57                                                                     & \yes                                                                        & \pg                            & \yes                         &                              &                              &                              &                              &                              \\
\multicolumn{1}{l|}{}                                          & \flsea                                                         & Stereo                                           & -                             & 19,596                                                                 & \yes                                                                        & \pg                            & \yes                         &                              & \yes                         &                              &                              &                              \\
\multicolumn{1}{l|}{}                                          & \sotrue                                                        & Stereo                                           & -                             & 8,497                                                                  & \yes                                                                        & \enc                          & \yes                         &                              & \yes                         &                              &                              &                              \\
\multicolumn{1}{l|}{}                                          & \uwStereo                                                      & \sim                                             & -                             & 29,568                                                                 & \no                                                                         & \sim                           & \yes                         & \yes                         &                              &                              &                              &                              \\
\multicolumn{1}{l|}{}                                          & \stereoAdapter                                                 & \sim                                             & Both                          & 40,000                                                                 & \yes                                                                        & \sim                           & \yes                         & \yes                         & \yes                         &                              &                              &                              \\
\multicolumn{1}{l|}{}                                          & \cellcolor[HTML]{ECF4FF}\textbf{UWSim (Ours)}                  & \cellcolor[HTML]{ECF4FF}\sim                     & \cellcolor[HTML]{ECF4FF}-     & \cellcolor[HTML]{ECF4FF}105,600                                        & \cellcolor[HTML]{ECF4FF}\yes*                                               & \cellcolor[HTML]{ECF4FF}\sim   & \cellcolor[HTML]{ECF4FF}\yes & \cellcolor[HTML]{ECF4FF}\yes & \cellcolor[HTML]{ECF4FF}\yes & \cellcolor[HTML]{ECF4FF}     & \cellcolor[HTML]{ECF4FF}     & \cellcolor[HTML]{ECF4FF}     \\
\multicolumn{1}{l|}{}                                          & \cellcolor[HTML]{ECF4FF}                                       & \cellcolor[HTML]{ECF4FF}                         & \cellcolor[HTML]{ECF4FF}Train & \cellcolor[HTML]{ECF4FF}19,950                                         & \cellcolor[HTML]{ECF4FF}                                                    & \cellcolor[HTML]{ECF4FF}-      & \cellcolor[HTML]{ECF4FF}\yes & \cellcolor[HTML]{ECF4FF}\yes & \cellcolor[HTML]{ECF4FF}\yes & \cellcolor[HTML]{ECF4FF}     & \cellcolor[HTML]{ECF4FF}     & \cellcolor[HTML]{ECF4FF}     \\
\multicolumn{1}{l|}{\multirow{-10}{*}{\rotatebox{90}{Stereo}}} & \multirow{-2}{*}{\cellcolor[HTML]{ECF4FF}\textbf{SUDS (Ours)}} & \multirow{-2}{*}{\cellcolor[HTML]{ECF4FF}Stereo} & \cellcolor[HTML]{ECF4FF}Test  & \cellcolor[HTML]{ECF4FF}1,486                                          & \multirow{-2}{*}{\cellcolor[HTML]{ECF4FF}\yes*}                             & \cellcolor[HTML]{ECF4FF}\pg    & \cellcolor[HTML]{ECF4FF}\yes & \cellcolor[HTML]{ECF4FF}\yes & \cellcolor[HTML]{ECF4FF}\yes & \cellcolor[HTML]{ECF4FF}     & \cellcolor[HTML]{ECF4FF}     & \cellcolor[HTML]{ECF4FF}     \\ \hline
\multicolumn{1}{l|}{}                                          & \uwslam                                                        & Stereo                                           & -                             & 3,111                                                                  & \yes                                                                        & \pg                            & \yes                         &                              &                              &                              &                              &                              \\
\multicolumn{1}{l|}{}                                          & \flsea                                                         & Mono.                                            & -                             & 22,456                                                                 & \yes                                                                        & \pg                            & \yes                         &                              & \yes                         &                              & \yes                         &                              \\
\multicolumn{1}{l|}{}                                          & \svin                                                          & Custom                                           & -                             & 25,145                                                                 & \yes                                                                        & \pg                            & \yes                         & \yes                         & \yes                         &                              & \yes                         & \yes                         \\
\multicolumn{1}{l|}{}                                          & \tank                                                          & Stereo                                           & -                             & 40,649                                                                 & \yes                                                                        & \at                            & \yes                         & \yes                         & \yes                         & \yes                         & \yes                         & \yes                         \\
\multicolumn{1}{l|}{\multirow{-5}{*}{\rotatebox{90}{SLAM}}}    & \cellcolor[HTML]{ECF4FF}\textbf{SUDS (Ours)}                   & \cellcolor[HTML]{ECF4FF}Stereo                   & \cellcolor[HTML]{ECF4FF}-     & \cellcolor[HTML]{ECF4FF}29,319                                         & \cellcolor[HTML]{ECF4FF}\yes*                                                & \cellcolor[HTML]{ECF4FF}\pg    & \cellcolor[HTML]{ECF4FF}\yes & \cellcolor[HTML]{ECF4FF}\yes & \cellcolor[HTML]{ECF4FF}\yes & \cellcolor[HTML]{ECF4FF}\yes & \cellcolor[HTML]{ECF4FF}\yes & \cellcolor[HTML]{ECF4FF}\yes \\ \hline
\end{tabular}
\end{table*}

\begin{figure}[h!tp]
    \centering
    \includegraphics[width=0.95\linewidth]{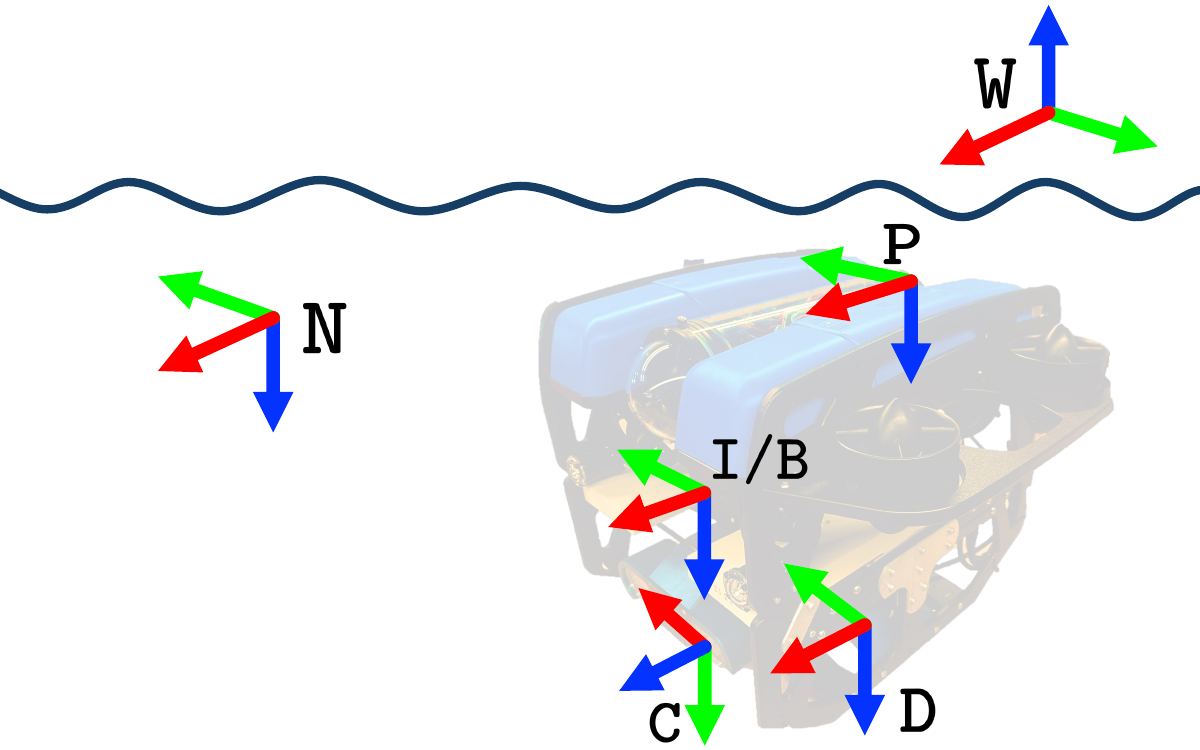}
    \caption{Coordinate frames of the robot. Red indicates the $ x$-axis, green the $ y$-axis, and blue the $ z$-axis. The IMU ($\Imu$) and base ($\Base$) frames are coincident. The DVL ($\DVL$) and barometer ($\Barometer$) frames share the same frame convention and the camera frame ($\Camera$) follows the OpenCV frame convention~\cite{noauthor_opencv_nodate}.}
    \label{fig:frames}
\end{figure}

\section{Underwater Stereo Estimation}\label{sec:stereo_depth}
This section introduces our proposed method for learning-based underwater stereo depth estimation, including data generation, data augmentations, implementation details, and training loss.
\subsection{Simulated Data Generation}\label{sec:ocean-sim}
Motivated by the lack of real-world underwater stereo datasets with ground truth depth, and inspired by the success of stereo disparity estimation in \cite{wen2025foundationstereo}, we leverage simulated data for network training. 
To generate training data, we first leverage the Omniverse RTX renderer \cite{Nvidia_omniverse} to generate rendered images in an IsaacSim environment \cite{isaacsim2025}, as shown in the first column of \Figure\ref{fig:augmentations}.
Objects and scene assets are collected from Epic Fab \cite{epic_fab} and Sketchfab \cite{sketchfab}, totaling eight scenes. 
These scenes are selected for their similarity in terrain properties (i.e., rocky and sandy seafloors) and structure (i.e., open water, caves).
The assets are selected to reflect common underwater structures, including large and small shipwrecks, rocks spanning a range of scales and textures, and marine vegetation.
During the initial simulated data generation stage, the position of the stereo camera pair is randomly sampled from a pre-recorded trajectory in each scene, and a random orientation at the position is applied to ensure diversity of views of the scene. 
With each render, the stereo pair image is generated, along with the camera intrinsics, depth image, and scene normals.

We note that when rendering the initial simulated images, we do not apply any underwater effects.
While there has been a recent rise in GPU-accelerated simulators for the underwater domain \cite{Potokar22icra, romrell2025previewholoocean20, song2025oceansim}, simulated effects are limited to haze and attenuation. We also seek to model caustics, dynamic lighting, and suspended particles. Thus, we propose an augmentation pipeline that takes in simulated in-air images and applies train-time augmentations. This has the added benefit of randomizing augmentations at training time, yielding multiple configurations of underwater image parameters for each simulated image.

\subsection{Underwater Data Augmentation}\label{sec:augmentations}
This section presents a training-time augmentation pipeline that transforms the simulated in-air stereo imagery into physically plausible underwater renderings while preserving dense disparity ground truth.
We explicitly model dominant components of underwater image formation and apply them as composable operators to the simulated dataset. Concretely, the pipeline comprises four augmentation modules: (1) ambient water-column illumination, (2) surface-induced caustics from refracted sunlight, (3) directional and diffuse sunlight, and (4) specular highlights from suspended sediment particles.
The cumulative effect of these operators on the input images is illustrated in \Figure \ref{fig:augmentations}.
Note that since the presented augmentations only require the simulated in-air stereo images, ground-truth depths, the camera intrinsics, and the stereo extrinsics, it is broadly transferable to other in-air datasets that contain dense ground-truth depth.

\subsubsection{Water Column Effects}
We model water-column effects using the light transport model described in~\cite{1593804,sethuraman2023waternerf}.
For an in-air image, $J_c$, with depth map, $\b{z}$, the simulated underwater pixel intensity, $I_c$, at location $\mathbf{x} = (u,v)$ is given by
\begin{gather}\label{eq:water-column}
    I_c(\b{x}, \b{z}) = J_c(\b{x})\cdot t_c(\b{z}) + B_{c,\infty} \cdot (1-t_c(\b{z}))\\
    t_c(\b{z}) = \exp\left(-\beta_c \b{z}\right) \label{eq:transmission}
\end{gather}
where $\beta_c$ is a per-channel attenuation coefficient for $c \in \{R,G,B\}$ and $B_{c,\infty}$ is the per-channel veiling-light (backscatter) term at infinite range.
We randomly sample $\beta_c$ according to Jerlov water types~\cite{jerlov1968optical}, and we sample $B_{c,\infty}$ over a range of conditions to span diverse underwater appearances at each time we apply the augmentation.
This approach is flexible enough to sample and generate the appearance of additional water column parameters for targeted augmentations to emulate various water conditions extending beyond the selected Jerlov water types~\cite{aquafuse}.
The effect of this model applied to a simulated in-air image is shown in the second column of \Figure \ref{fig:augmentations}.\\

\subsubsection{Caustics}
Caustics arise from sunlight refraction at the dynamic water surface \cite{upstill-caustics}; while analytic, surface-based caustic rendering is available in real-time graphics engines \cite{song2025oceansim}, it is impractical to integrate such pipelines within a train-time augmentation.
We instead model underwater caustics via texture mapping.
Given an image $J_c$ with depth map $\b{z}\left(\b{x}\right)$, we compute per pixel scene normals, $\b{N}\left(\b{x}\right)$, and apply a caustic texture map, $T_\text{caustic}$, under a planar light source aligned with the scene $+z$-axis with direction $\b{L}_z$.
The resulting caustic intensity image is
\begin{equation}
    I_\text{caustic}(\mathbf{x}) = T_\text{caustic}(\mathbf{x}) \max(0, \langle\mathbf{L}_z, \mathbf{N}(\mathbf{x}) \rangle ).
\end{equation}
This single-channel image is used to relight the image $J_c$.
For stereo pairs, we use the known stereo extrinsics to warp the texture map for consistent caustic projection.
The effect of the caustic simulation is shown in the third column in \Figure\ref{fig:augmentations}.
In practice, the normals are computed using Kaolin~\cite{jatavallabhula2019kaolinpytorchlibraryaccelerating}.
\begin{figure*}[h!tp]
    \centering
    \includegraphics[width=\linewidth]{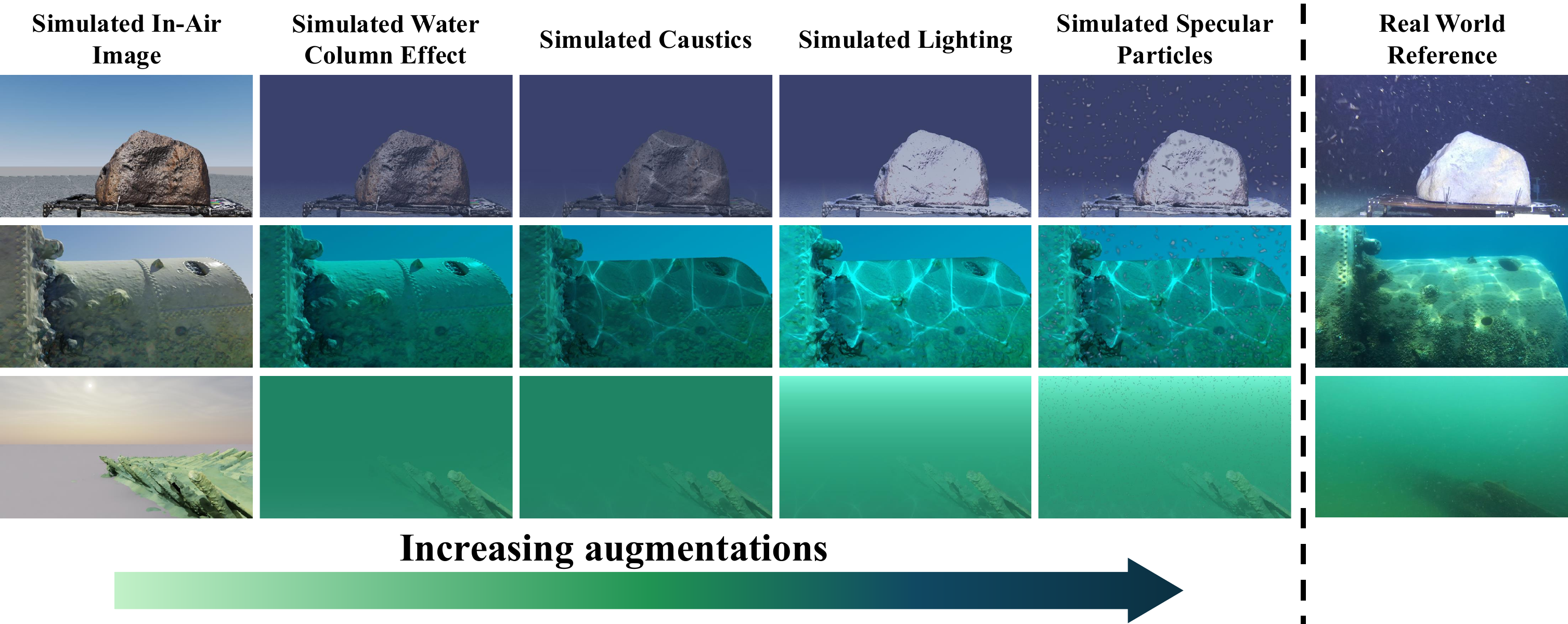}
    \caption{A demonstration of our underwater augmentation pipeline. From left to right, the first five columns demonstrate increasing levels of augmentation. The rightmost column shows the reference image that the augmentations aim to match.}
    \label{fig:augmentations}
\end{figure*}
\subsubsection{Directional Light}
Inspired by the presence of directional lighting in underwater datasets (see Table~\ref{tab:datasets}) and by methods that explicitly exploit lighting for reconstruction \cite{roznere_lights}, we incorporate directional light simulation into our dataset generation.

Given the depth map, $\b{z}\left(\b{x}\right)$, and its computed normals, $\b{N}\left(\b{x}\right)$, we place a point light source at position $p$ in the camera frame with pose $P$ and color $C_c$.
For a pixel at location $\b{x} = \left(u, v \right)$, the light direction is
\begin{gather}
    \mathbf{L}(\mathbf{x}) = p - \b{z}\left(\b{x}\right) K^{-1}\begin{bmatrix}
        u & v & 1
    \end{bmatrix}^T.
\end{gather}
where $K$ is the camera intrinsic matrix. The contribution of the directional light to the original image is
\begin{gather}
    I_\text{light,c}(\mathbf{x}) = \frac{PC_c}{\lVert\mathbf{L}(\mathbf{x})\rVert^2}\max\left(0,\langle \mathbf{L}(\mathbf{x}), \mathbf{N}\rangle\right) t_c(\lVert\mathbf{L}(\mathbf{x})\rVert),
\end{gather}
where $t_c$ is depth-dependent, per-channel attenuation defined in \eqref{eq:transmission}.
For a stereo pair, we keep $p$ fixed in the left camera frame and simulate the light contribution on the right image.
The effect of directional lighting is shown in the fourth column in \Figure\ref{fig:augmentations}.\\

\subsubsection{Sunlight}
In shallow-water applications, sunlight significantly affects the captured image.
This is evident in the real-world image shown in the bottom row of \Figure\ref{fig:augmentations}.
Effects like crepuscular rays are prevalent in these environments, and the strong sunlight source contributes to a gradient of color change in the water column~\cite{godrays}. 
We model two dominant sunlight modes purely in image space, without relying on the 3D scene geometry: planar and radial illumination. 

For planar illumination, we impose a vertical intensity gradient in the image, with pixels near the top receiving the strongest contribution.
This is parametrized by the sun color, $C_\text{sun}$, the additional brightness, $b_\text{planar}$, and saturation distance in image space, $d_\text{sat}$.
Color and saturation changes are computed consistently with the water column parameters, $\beta_c$ and $B_\infty$, used to add the water column effects.
For radial illumination, we define $C_\text{sun}$ and a sun position, $p_\text{sun}$, in image coordinates; radial distance from $p_\text{sun}$ controls the color and intensity change, again tied to $\beta_c$ and $B_\infty$. 
Sampling $p_\text{sun}$ above the image plane increases the curvature of the halo, mimicking a sun just outside the camera’s field of view.
The effect of a planar sunlight on the augmented image is shown in the fourth column in \Figure\ref{fig:augmentations}.\\

\subsubsection{Particles}
Suspended particles are a common feature of underwater imagery, arising from high turbidity or sediment in the water column.
We model only particles located between the scene and the camera. 
To simulate particle appearance, we apply a procedurally generated particle texture map directly in image space. For each particle texture map, we randomly place a set of ellipsoids across the image, varying in orientation and size.
Each ellipsoid's opacity decreases radially outward from its center, simulating the motion blur experienced during image capture.
The resulting particle texture map is additively blended with the base image.
The effect of simulated particles is shown in the fifth row of \Figure\ref{fig:augmentations}.\\

\subsubsection{Full Augmentation Pipeline}
For the full augmentation pipeline, shown graphically in Figure~\ref{fig:augmentations}, we apply all lighting effects first, then add water haze to simulate an underwater image.
We define a relit image $I_{\text{relit},c}$ as
\begin{align}
    I_{\text{relit},c}(\mathbf{x}) = J_c(\mathbf{x})\left(1+I_{\text{light},c}(\mathbf{x})\right)\left(1+I_\text{caustic}(\mathbf{x})\right).
\end{align}

We apply the water-column model to $I_{\text{relit},c}$ to ensure all illumination is subject to medium effects:
\begin{gather}
    I_{\text{water}, c}(\mathbf{x}, \mathbf{z}) = I_{\text{relit},c}t_c(\mathbf{z})+B_\infty(1-t_c(\mathbf{z})).
\end{gather}

As a final step, we apply the sunlight and particles directly onto $I_\text{water}$.
The parameters of each augmentation are randomly sampled to form a large distribution of water effects.

\subsection{Stereo Training}
In this section, we describe a unified training strategy for fine-tuning existing stereo depth networks in underwater environments.
We combine supervised objectives on our augmented synthetic data with self-supervised objectives on real underwater stereo data and apply this strategy across a range of model sizes and architectures~\cite{defom_stereo, wen2025foundationstereo}.

\subsubsection{Loss Functions}
Stereo networks produce a set of refined disparity estimates $\hat{\b{d}} = ( \hat{d}_0, \dots, \hat{d}_R )$, where $\hat{d}_r$ is the final prediction after $R$ refinement steps.
The training loss switches between supervised ($\L_\text{sup}$) and self-supervised ($\L_\text{self})$ losses depending on whether the ground-truth disparity, $d^*$, is available. That is,
\begin{equation}
    \L(\disp) = \begin{cases}
        \L_\text{sup}(\disp, d^*) & d^* \text{ known} \\
        \lambda_\text{self} \cdot \L_\text{self}(\hat{d}_R) & \text{ otherwise}
    \end{cases}
\end{equation}
where $\lambda_{\text{self}}$ is a scalar hyperparameter.

\subsubsection{Direct Disparity Supervision}
Ground-truth disparity, $d^*$, is available only for simulated data.
For training on simulated data, to avoid supervising on regions of the image severely degraded by water column effects, we use the known per-channel attenuation $t_c(\b{x})=\exp(-\beta_c\b{x})$ from our augmentation pipeline.
We mask out regions where $t_c(\b{x}) < t_{min, c}$ for any channel, retaining only pixels that carry information about the scene geometry.
\begin{equation}
d^*(\mathbf{x}) = d^*(\mathbf{x}) \, \mathbb{1}\!\left[ t_c(\mathbf{x}) \ge t_{min,c} \right]
\end{equation}
With this masked ground-truth, we use the same loss function as in \cite{wen2025foundationstereo}:
\begin{align}\label{eq:loss-supervised}
\begin{split}
    \mathcal{L}_\text{sup}(\disp, d^*) = \lvert \hat{d}_0 - d^*\rvert_\text{smooth} + \sum_{r=1}^R\gamma^{R-r}\lVert \hat{d}_r - d^*\rVert_1
\end{split}
\end{align}
where $\lvert\cdot\rvert_\text{smooth}$ denotes the Smooth L1 loss and $\lVert\cdot\rVert_1$ denotes the standard L1 Loss, as described by \cite{girshick2015fastrcnn, wen2025foundationstereo}.\\

\subsubsection{Self-Supervised Warping Loss}\label{sec:self-supervision}
When ground-truth disparity is not available, we use an alternate supervision strategy.
Following several past works \cite{vankadari2024dtd, skinner2019uwstereonet}, we use a self-supervising loss that uses the estimated disparity to warp each image to the frame of the other.
That is, given a rectified pair of images $\left(I_L, I_R\right)$ along with the rectified intrinsics, $K$, and the stereo baseline, $b$, each image is back-projected to 3D by converting the disparities to depths.
Then, the 3D points are projected into the frame of the other image and interpolated to form the final image.
We denote this operation $\hat{I}_L= \texttt{warp}(I_R, d)$ and $\hat{I}_R = \texttt{warp}(I_L, d)$, where the dependence on calibrations is omitted for brevity.
In practice, this warp operation is implemented by the open-source Kornia library \cite{eriba2019kornia}.

Using the estimated images and ground-truth, the following loss is computed:
\begin{gather}
\hat{I}_L = \texttt{warp}(I_R, \hat{d}_K)\hspace{16pt} \hat{I}_R = \texttt{warp}(I_L, \hat{d}_K)\\
 \L_\text{warp}(\hat{d}_K) = \frac{1}{2}\bigg[\L_\text{DTD}(\hat{I}_L, I_L)+ \L_\text{DTD}(\hat{I}_R, I_R)\bigg], \label{eq:warp-loss}
\end{gather}
where
\begin{align}
\begin{split}
\L_\text{DTD}(I, \hat{I}) = &\lambda_\text{DTD}\cdot\lvert I - \hat{I}\rvert_1 \\
+ &(1-\lambda_\text{DTD})\cdot\text{SSIM}(I, \hat{I})
\end{split}
\end{align}
is the loss function proposed by \cite{vankadari2024dtd}. Unlike the directly supervised loss in \eqref{eq:loss-supervised}, the self-supervised loss is applied only to the final upsampled prediction $\hat{d}_K$.

While \cite{vankadari2024dtd} applies warping only from the right image to the left, we follow \cite{skinner2019uwstereonet} and apply the loss in both directions. 
Although \eqref{eq:warp-loss} is written as the average of the two losses, in practice, the warped image may contain invalid pixels at the edges (caused by parts of the scene being out of view in one camera).
In such cases, we keep only the valid camera’s loss and weight it twice, while discarding the invalid camera’s loss.\\

\subsubsection{Occam Regularizer}
Even though \eqref{eq:warp-loss} provides effective self-supervision in textured regions, low-texture areas cause many disparities to produce the same warped image, especially in background regions where disparity should be zero.
To resolve this, we add an \textit{Occam regularizer} that favors zero disparity when it yields results similar to larger disparities.

Let $\L_\text{warp}(\hat{d}_K)$ be the photometric loss in \eqref{eq:warp-loss} for the predicted disparity map $\hat{d}_K$.
We compute an alternative loss $\L_\text{warp}(0)$ corresponding to a \emph{zero-disparity hypothesis} by directly comparing the unwarped stereo pair.
Then, we define
\begin{align}
\Delta_\text{occam}(\hat{d}_K) = \L_\text{warp}(\hat{d}_K) - \L_\text{warp}(0),
\end{align}
which measures the gain from the predicted disparity over zero disparity.

If the prediction is significantly worse than zero disparity, we add a penalty:
\begin{align}
\L_\text{occam}(\hat{d}_K) = \texttt{ReLU}(\tau_\text{occam} + \Delta_\text{occam}(\hat{d}_K)),
\end{align}
where $\tau_{occam}$ is a margin parameter.\\

\subsubsection{Total Loss Function}Finally, as in \cite{vankadari2024dtd}, we add the edge-aware smoothness regularizer, $\L_\text{smooth}$, from \cite{godardmonodepth2017}. Thus, the final self-supervised loss is
\begin{align}
\begin{split}
\mathcal{L}_\text{self}(\hat{d}_K) = 
&\lambda_\text{warp} \mathcal{L}_\text{warp}(\hat{d}_K)  \\
+ &\lambda_\text{occam}\mathcal{L}_\text{occam}(\hat{d}_K) \\
+ &\lambda_\text{smooth}\mathcal{L}_\text{smooth}(\hat{d}_K)
\end{split}
\end{align}
where $\lambda_\text{warp}, \lambda_\text{occam}$ and $\lambda_\text{smooth}$ are scalar weights determined experimentally.
In practice, we train the model with two stages.
In the first stage, we train using only simulated data for a fixed number of iterations.
We then fine-tune on the second stage using a mixture of simulated and real-world data.
\section{SurfSLAM}\label{sec:slam}
Our proposed algorithm \algname applies the finetuned stereo depth estimation module and tightly integrates the IMU, DVL, barometer, and stereo depth maps to estimate the robot's pose and to build a dense 3D reconstruction of the scene. 
We base our tracking system on TURTLMap \cite{song2024turtlmap} due to its metrically consistent odometry in low-texture underwater environments.
We treat the stereo-derived geometry as a high-level measurement used to correct drift with global registration factors while relying on acoustic–inertial sensing for locally consistent motion.
For the sake of clarity, we describe the full acoustic-inertial-stereo method below.

\subsection{State Definition}
The robot state at a keyframe $i$ is represented as a combination of the pose, velocity, and sensor biases:
\begin{equation}
    \mathbf{x}_i = \left[\mathbf{R}_i, \mathbf{p}_i, \mathbf{v}_i, \mathbf{b}_i^g, \mathbf{b}_i^a, \mathbf{b}_i^v\right] \in \SOthree \times \R^{15},
\end{equation}
where $\mathbf{R}_i \in \text{SO}(3)$ and $\mathbf{p}_i \in \R^3$ represent the orientation and position in the NED frame, $\mathbf{v}_i\in \R^3$ represents the body frame linear velocity, $\b{b}_i^g, \b{b}_i^a \in \R^3$ are the IMU gyroscope and accelerometer biases, and $\mathbf{b}_i^v \in \R^3$ is the DVL velocity bias.
We maintain the states over a set of keyframes
\begin{equation}
    \cal{X}_n = \{\mathbf{x}_i\}_{i\in\mathsf{K}_n},
\end{equation}
where $\mathsf{K}_n$ represents the keyframes up until time $t_n$.
In practice, we add keyframes at 1 Hz to balance factor graph complexity, stereo inference time, and estimation accuracy.

\subsection{Measurement Definitions}
We consider sensor measurements from an IMU, a DVL, and a barometer between two consecutive keyframes $i$ and $j$.
For a stereo camera, we consider the measurements between a query keyframe at time $q$ and a matched keyframe at time $m$.

The depth estimates from the stereo camera are denoted $\cal{S}_{ij}$.
The IMU measurements are denoted as $\cal{I}_{ij}$, and are composed of a set of angular velocity $\tilde{\mathbf{\omega}}\in \R^3$ and linear acceleration $\tilde{\mathbf{a}}\in \R^3$ measurements. 
The DVL measurements, denoted $\cal{D}_{ij}$, consist of a linear velocity $\tilde{\mathbf{v}} \in \R^3$ measurement, and the barometer measurement, denoted as $P_{ij}$ is composed of a pressure measurement, $\tilde{z}$.
The full set of measurements is denoted
\begin{equation}
    \cal{Z}_n = \{\cal{S}_{qm}, \cal{I}_{ij}, \cal{D}_{ij}, \cal{P}_{ij}\}_{i,j,q,m \in \mathsf{K}_n}.
\end{equation}

\label{subsec:acoustic-inertial}
\subsection{Acoustic-Inertial Pose Graph Formulation}
We follow the formulation presented in~\cite{song2024turtlmap} for acoustic-inertial odometry.
We denote the subset of acoustic-inertial measurements as $\underline{\cal{Z}}_k$ consisting of the IMU, DVL and barometer until time $t_k$:
\begin{equation}
    \underline{\cal{Z}}_k = \{\cal{I}_{ij}, \cal{D}_{ij}, \cal{P}_{ij}\}_{i,j \in \mathsf{K}_k}.
\end{equation}
The estimator optimizes state estimates to maximize the posterior distribution given the measurements $\underline{\cal{Z}}_k$ and a prior distribution over the states $\cal{X}_0$:
\begin{align}
    \cal{X}_k^* &= \argmax_{\cal{X}_k} p\left(\cal{X}_k \vert \underline{\cal{Z}}_k\right) \\
     &\propto \argmax_{\cal{X}_k} p\left(\cal{X}_0\right) p(\underline{\cal{Z}}_k \vert \cal{X}_k)
\end{align}
We assume that the sensor measurements are conditionally independent and have zero-mean Gaussian noise.
Following this, we can transform the objective function into a weighted least squares problem by minimizing the log-likelihood of each measurement and the prior over the states, weighted by their associated uncertainties.
This objective is given as

\begin{align}
    \cal{X}^*_k = &\argmin_{\cal{X}_k} \lVert \b{r}_0 \rVert^2_{\Sigma_0} \nonumber \\ &+ \sum_{i,j \in \mathsf{K}_k}\left( \lVert \b{r}_{\cal{I}_{ij}} \rVert^2_{\Sigma_{\cal{I}_{ij}}} + \lVert \b{r}_{\cal{D}_{ij}} \rVert^2_{\Sigma_{\cal{D}_{ij}}} +  \lVert \b{r}_{\cal{P}_{ij}} \rVert^2_{\Sigma_{\cal{P}_{ij}}}\right),
\end{align}
where $\lVert \cdot \rVert^2_\Sigma$ denotes the Mahalanobis distance with covariance $\Sigma$, and $\b{r}$ denotes the residual associated with the factor for the measurements $\underline{\cal{Z}}_k$ and the prior belief $\cal{X}_0$.
In practice, this optimizer is solved iteratively as measurements arrive using iSAM2 \cite{kaess2012isam2}.
The residuals and noise models for the sensors are derived in~\cite{10149804, Forster_2017, song2024turtlmap}.

\subsection{Global Registration}
This section describes how the tracking method from \cite{song2024turtlmap} is augmented with global registration. 
Global registration is performed in three steps: (1) global place recognition, (2) visual feature matching, and (3) geometric alignment. Each step is described below.

\subsubsection{Global Place Recognition}
Let $\mathsf{K}_q$ denote the set of keyframes up to a time $t_q \leq t_k$, and let $k_q \in \mathsf{K}_q$ be the $q^{th}$ keyframe that is used to query the place recognition system.
The left image of $k_q$, denoted $I_{q,l}$, is first histogram equalized using Contrast-Limited Adaptive Histogram Equalization (CLAHE)~\cite{clahe}. Then, SuperPoint features (as proposed and trained by \cite{detone2018superpoint, sarlin2020superglue}) are extracted from the equalized image.
The local feature descriptors from the SuperPoint features are accumulated into an image-level descriptor using the Vector of Locally Aggregated Descriptors (VLAD) method \cite{jegou2010vlad}.
In practice, the K-nearest neighbors process within VLAD is implemented via \cite{douze2025faiss}.
The image-level descriptors computed by VLAD are compared with other keyframes that are at least 20 seconds old.
The top three most similar images are retrieved and passed to the visual feature matching step.

\subsubsection{Visual Feature Matching}
We denote a matched image candidate as $I_{m,l}$.
Initial feature matching is performed using SuperGlue \cite{sarlin2020superglue}.
Outliers are rejected by computing the essential matrix between the query and match images using RANSAC.
If RANSAC estimates sufficient inliers, the match candidate proceeds to the geometric validation step. 

\subsubsection{Geometric Validation}
Using the stereo estimation described in \ref{sec:stereo_depth}, a depth image is computed for both the query and matched images.
Using these depth images, the filtered 2D correspondences estimated by SuperGlue are lifted to 3D.
These 3D feature correspondences are used to estimate a 3D transformation, $\b{T}_{qm}^C$, between the two keyframes.
This is again performed using RANSAC to reject poor correspondences. If insufficient inliers are detected, the registration is rejected.
Otherwise, the RANSAC-estimated transform is refined using GICP \cite{confsegal2009gicp} on the full point clouds estimated by the stereo depth estimation network, yielding a final geometric relative pose measurement $\tilde{\b{T}}_{qm}^C \in \SEthree$, which we simply denote as the stereo measurement, $\cal{S}_{qm}$.

While global place recognition produces three candidate frames, at most one is passed to the optimizer. That is, the first match candidate to survive the above heuristics is passed to the pose graph, while any remaining candidates are discarded.

\subsubsection{Pose Graph Formulation}
We incorporate $\cal{S}_{qm}$ as a factor added to the acoustic-inertial tracking to constrain the relative motion between the keyframes.
Before adding the registration to the pose graph, a final probabilistic test is performed.
A stereo factor, $\cal{S}_{qm}$, is added to the graph after keyframes at time $t_q$ and $t_m$ have been optimized using the acoustic-inertial factors described in Section \ref{subsec:acoustic-inertial}.
As a result, the tracking system can be queried for a belief over $T^C_{qm}$, including mean $\mu_{T^C_{qm}}$ and marginal covariance $\Sigma_{T^C_{qm}}$.
We compute the Mahalanobis distance between the stereo registration result and the tracker's estimate according to 

\begin{equation}
d_{qm} = 
\left\| \, T^C_{qm} - \mu_{T^C_{qm}} \right\|_{\Sigma_{T^C_{qm}}}^2
\end{equation}
Registrations with $d_{qm} > 2.5$ are treated as outliers and discarded; only those satisfying $d_{qm} \le 2.5$ are added to the pose graph as stereo factors $\cal{S}_{qm}$.

In practice, each of the outlier rejection methods is tuned to add as many registrations to the factor graph as possible. 
To deal with the noisy estimates resulting from this strategy, registration factors are weighted by a robust Huber kernel~\cite{huber1992robust} in the backend to reject incorrect registrations that survive the outlier rejection process.

The final objective becomes
\begin{align}
    \cal{X}^*_k = &\argmin_{\cal{X}_k} \lVert \b{r}_0 \rVert^2_{\Sigma_0} \nonumber \\ &+ \sum_{i,j \in \mathsf{K}_k}\left( \lVert \b{r}_{\cal{I}_{ij}} \rVert^2_{\Sigma_{\cal{I}_{ij}}} + \lVert \b{r}_{\cal{D}_{ij}} \rVert^2_{\Sigma_{\cal{D}_{ij}}} +  \lVert \b{r}_{\cal{P}_{ij}} \rVert^2_{\Sigma_{\cal{P}_{ij}}}\right) \\
    &+ \sum_{q,m\in\mathsf{K}_q}\rho\left(\lVert\b{r}_{\cal{S}_{qm}}\rVert^2_{\Sigma_{\cal{S}_{qm}}}\right),
\end{align}
where the stereo pose registration residual $\b{r}_{\cal{S}_{qm}}$ is defined as
\begin{equation}
    \b{r}_{\cal{S}_{qm}} = \text{Log}\left(\tilde{T}_{qm}^{-1}\bar{T}_{qm}\right) \in \mathfrak{se}(3)
\end{equation}
where $\tilde{T}_{qm}$ represents the pose measurement from the GICP estimate, $\bar{T}_{qm}$ represents the estimated relative pose, $\text{Log}$ represents the mapping from the Lie group $\SEthree$ to its Lie algebra $\mathfrak{se}(3)$ and $\rho$ denotes the robust Huber kernel.

\subsection{Mapping}
For mapping, we adopt a similar approach to that of~\cite{teed2021droid} and maintain the stereo depth at each keyframe as a map frame primitive. 
Each incoming keyframe contributes a depth image that is back-projected into a local point cloud using the calibrated stereo intrinsics and extrinsics.
These keyframe point sets are accumulated to form an explicit geometric reconstruction of the environment.

To address noise introduced to the pointcloud by depth discontinuities, we apply a lightweight, GPU-based outlier rejection and noise cleaning step prior to the map frame construction.
Specifically, we apply an edge cleaning, followed by a statistical outlier removal and radius outlier removal step. 
This results in a map with minimal noise and errors and maintains the global registrations that anchor the pose during the tracking phase of \algname.
\section{Experiments}\label{sec:experiments}

\subsection{Data Collection and Generation}
Real-world underwater stereo data was collected at the Thunder Bay National Marine Sanctuary in Alpena, MI, with multiple shipwreck sites and in an outdoor test tank across multiple days. 
The data collection platform is a human-operated BlueROV2 with a forward-facing stereo camera.
Field sequences span depths from 4 to 20 m and rely on natural illumination; additional sequences were acquired around an artificial rock target in an outdoor pool (both day and night), where the BlueROV2 lights provide additional lighting as needed. 
The dataset exhibits a broad range of turbidity, caustics, lighting, and suspended particulates (see the rightmost column of ~\Figure~\ref{fig:augmentations} and leftmost column of \Figure~\ref{fig:gt_annotations}).

\begin{figure}
    \centering
    \includegraphics[width=1.0\linewidth]{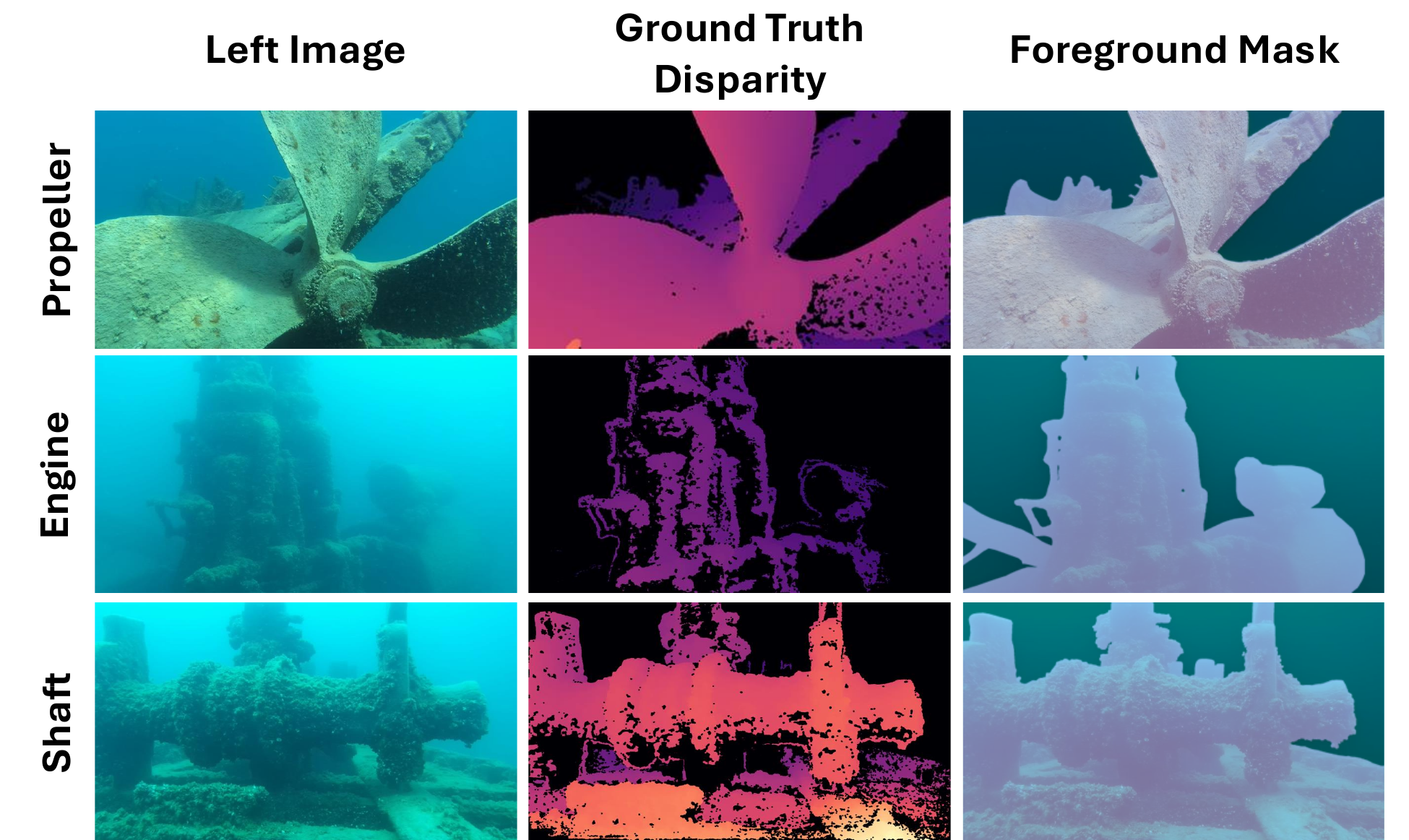}
    \caption{An example from three of the nine stereo sequences is shown. The first column shows the left rectified image; the second column shows the disparity estimated from our metric photogrammetry pipeline; the last column shows the manually-annotated foreground mask overlaid on the left image. The disparity color maps are so that lighter (yellow) is higher disparity and darker (purple) is lower disparity, with black being zero disparity.}
    \label{fig:gt_annotations}
\end{figure}

\begin{figure*}[!th]
    \centering
    \includegraphics[width=1.0\linewidth]{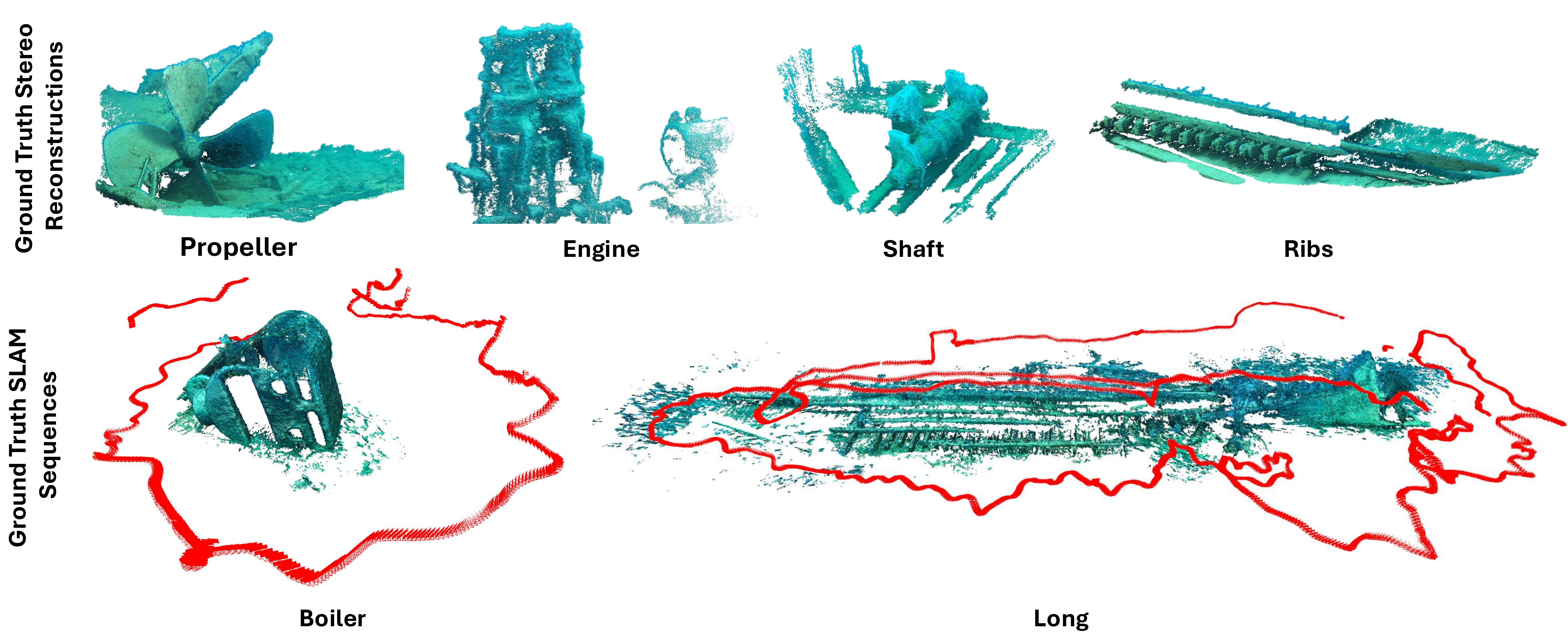}
    \caption{A visualization of a subset of the ground truth metric reconstructions from the stereo (first row) and SLAM sequences (second row) used for evaluations. 
    The SLAM sequences show the ground-truth poses as red camera frustums and a sparse COLMAP reconstruction.
    The Engine SLAM sequence, not pictured, is a second sequence around the right-hand side of the Long reconstruction.}
    \label{fig:reconstructions}
\end{figure*}

Our collected dataset, which we call \textbf{S}tereo \textbf{U}nderwater \textbf{D}ataset for \textbf{S}hipwrecks (\textbf{SUDS}), contains three subsets: (i) stereo evaluation, (ii) stereo training, and (iii) SLAM evaluation.
All three dataset splits were collected at different shipwreck sites. Test data was collected at shipwreck sites that were not seen during training.

Our simulation dataset, which we call \textbf{U}nder\textbf{{W}}ater \textbf{Sim}ulation (\textbf{UWSim}), contains simulated data of realistic underwater scenes generated as described in Section~\ref{sec:ocean-sim} and is used for training our stereo networks.
This differs from alternative datasets used in underwater stereo training, which use either simulated images from non-underwater images or lack the visual properties of underwater sequences.

SLAM evaluation comprises three sequences on a single shipwreck under varying haze, texture, and geometry. 
The dataset statistics are provided in \Table \ref{tab:datasets}.

\subsubsection{Real-world Stereo Dataset}
The real-world stereo training dataset consists of twenty-two real-world sequences without accompanying ground-truth data, intended for self-supervision, with detailed statistics in \Table\ref{tab:datasets}.
This dataset is composed of a variety of underwater lighting conditions, turbidity levels, and distractors.
In addition to the stereo images, this dataset also consists of the stereo intrinsic and extrinsic parameters, making the dataset suitable for self-supervised training, as described in Section~\ref{sec:self-supervision}.

The SUDS stereo evaluation dataset comprises nine sequences across five shipwreck sites, totaling 1,486 stereo pairs with labeled ground-truth disparity, poses, and 3D reconstruction.
The ground truth for the stereo evaluation dataset was obtained through a metric photogrammetry pipeline.
We provide stereo intrinsic and extrinsic parameters and per-frame depth maps through the metric photogrammetry pipeline.
Reconstructions of four sequences are shown in the first row of \Figure~\ref{fig:reconstructions}.

We note that a key evaluation metric is to assess the success of stereo depth estimation methods in the water column.
Stereo depth estimation methods trained in air tend to mistake the water column for scene geometry.
To assess the ability of stereo networks to reject the background, we manually annotated each image with foreground-background masks.
These manual annotations were augmented with automated ground plane annotations.
To reconstruct the ground plane, the distance to the ground plane was computed for each camera pose using the DVL's altitude measurements from its four beams.
A plane was fit to these ground measurements and projected into each camera view.
The foreground-background masks are used only for evaluation and reporting metrics, and are never used in training.

Sample outputs from the evaluation dataset are shown in~\Figure\ref{fig:gt_annotations}.
All stereo evaluations are conducted on our real-world dataset.

\subsubsection{Real-world SLAM Datasets}
With SUDS, we provide three sequences for evaluating \algname.
Each sequence includes 1080p stereo video at 30 fps with time-synchronized IMU, DVL, and barometer measurements.
The IMU is recorded at 200 Hz, the DVL at 8 Hz and the barometer at 5 Hz.

Ground truth is generated by taking a downsampled 5 fps version of the 1080p stereo video and running the metric photogrammetry pipeline, detailed in Section \ref{sss:metric_photo}.
This ensured that even in regions where there is low texture, the photogrammetry would track the camera poses and yield a near-full trajectory of the robot with minimal dropouts.
To the extent of our knowledge, the SLAM sequence of the SUDS dataset presents the first publicly released in-the-wild dataset with ground truth trajectory and maps for a robot equipped with an IMU, DVL, barometer, and stereo camera. 
We refer to these three scenes as \textit{Boiler}, \textit{Engine}, and \textit{Long}.
A reconstruction of the \textit{Boiler} and \textit{Long} scenes, along with the ground truth camera trajectories, is shown in the second row of \Figure\ref{fig:reconstructions}.

The \textit{Boiler} sequence is a 180-second-long trajectory circling a large boiler followed by a short traversal over it. 
The \textit{Engine} sequence is a 284-second-long run alongside the engine section of a shipwreck, transitioning from motion along the engine to lateral passes in front of the stern while inspecting the propeller.
Of the three sequences, the Engine sequence is the most consistently visually rich sequence.
The \textit{Long} sequence is a 550-second-long survey spanning long distances along the shipwreck, including extended intervals over low-texture regions observing the terrain.

\subsection{Generating Ground Truth for Real Data}
For data collected in the field, ground truth is generated using a metric photogrammetry pipeline.

\subsubsection{Metric Photogrammetry Pipeline} \label{sss:metric_photo}
We follow a two-step approach to obtain accurate metric reconstructions, camera poses, intrinsic calibration, and extrinsic calibration of both the stereo pair and IMU.
First, structure-from-motion (SfM) is run on sequences of uncalibrated stereo images using COLMAP~\cite{schoenberger2016sfm, schoenberger2016mvs}.
An approximate intrinsic calibration produced by Kalibr~\cite{kalibr} is used as an initial guess for COLMAP.
We then apply a custom optimizer to resolve the metric scale of the scene.
The formulation of this metric optimization differs between sequences used for SLAM and those used for stereo evaluation.

\subsubsection{Stereo-IMU SLAM Metric Optimization}
To recover the metric scale of the scene, we develop a custom optimizer.
Kalibr~\cite{kalibr} is first used to estimate initial camera intrinsics and stereo-IMU extrinsics.
Feature tracks are then exported from COLMAP and imported into a factor graph optimizer.
IMU data is recovered from data logs and temporally associated with the corresponding images.
Following an implementation closely based on~\cite{carlone2014eliminating}, map points are marginalized while camera poses, intrinsics, and extrinsics are jointly optimized.
This optimization is solved in batch over all SLAM sequences to enforce a consistent calibration.

After batch optimization, the intrinsic and extrinsic calibrations are fixed, and camera poses and map points are refined on a per-sequence basis.
During per-sequence refinement, map points are explicitly treated as optimization variables to allow robust kernels to downweight poor correspondences from COLMAP.

\subsubsection{Stereo-Only Stereo Metric Optimization}
For stereo-only sequences, no IMU data is available due to equipment failure during field testing.
As a result, the scene scale cannot be directly optimized when the stereo baseline is unknown.
In this case, the stereo baseline is fixed to the value estimated by Kalibr.
Without IMU factors, a two-stage refinement is unnecessary.
Instead, camera poses, intrinsics, stereo extrinsics, and map points are jointly optimized to minimize reprojection error.
This optimization is performed in batch across all stereo sequences.

\subsection{Stereo Training Details}
We provide details on the training done for fine-tuning existing stereo networks for underwater stereo depth prediction.

\subsubsection{Training Datasets}
\paragraph{Datasets For Supervised Training}
Supervised pretraining uses three simulated stereo datasets: UWSim (ours), TartanAir, and FlyingThings3D~\cite{wang2020tartanair, MIFDB16}.
We apply the full underwater augmentation pipeline only to UWSim and TartanAir, and keep FlyingThings3D unaugmented to preserve a reference domain.
This reduces the effective domain shift introduced by augmentations and helps prevent optimization instabilities that can otherwise lead to divergence and degraded accuracy in model performance.

\paragraph{Datasets for Self-Supervised Training}
For self-supervised training, we utilize real-world SUDS sequences in conjunction with additional in-the-wild underwater stereo data from SVIn2 and UWslam (Lizard Island), leveraging the provided calibrated stereo intrinsic and extrinsic parameters~\cite{svin2, lizard}. 
These sequences include broad viewpoint variation and diverse water and scene conditions, expanding the coverage of underwater appearance distribution used during the self-supervised finetuning.

\begin{table}[]
\centering
\small
\caption{Training parameters for the model finetuned for underwater stereo disparity estimation.}
\label{tab:configurations}
\setlength{\tabcolsep}{5pt}
\renewcommand{\arraystretch}{1.05}
\begin{tabular}{l|l||c}
\hline
Parameter & Description & Value \\ \hline \hline
Learning Rate           & Optimizer step size     & $1\mathrm{e}{-5}$ \\
$\lambda_\text{occam}$  & Occam loss weight       & 1.0 \\
$\tau_\text{occam}$     & Occam margin            & 0.01 \\
$\lambda_\text{smooth}$ & Smoothness weight       & 0.005 \\
$\lambda_\text{warp}$   & Photometric loss weight & 10.0 \\
$t_{\text{min},c}$      & Visibility threshold    & 0.05 \\
\hline
\end{tabular}
\normalsize
\end{table}

\useunder{\uline}{\ul}{}
\useunder{\uline}{\ul}{}
\begin{table*}[h!]
\centering
\footnotesize
\caption{We separate stereo evaluations into Large Models with ViT-L backbones and Small Models that use ViT-S or no vision transformer. In both cases, `Ours' refers to DEFOM with our simulation and self-supervised training. EPE (End Point Error) is measured in pixels. BP-1.0 and D1 are described in Section \ref{subsec:disp_metrics}. Lower is better for all metrics. The Params columns give the total number of parameters in each model and, separately, the number that are trainable, which discounts any frozen backbones.}
\begin{tabular}{|l|lcc||ccc|ccc|ccc|}
\hline
                       & \multicolumn{1}{c|}{Model} & \multicolumn{2}{c||}{Params (M)}             & \multicolumn{3}{c|}{Combined}                  & \multicolumn{3}{c|}{On Geometry Only}          & \multicolumn{3}{c|}{Water Column Only}           \\\cline{2-13}
                       & \multicolumn{1}{l|}{Method}                   & Total          & Train.         & EPE           & BP-1.0         & D1            & EPE           & BP-1.0         & D1            & EPE           & BP-1.0         & D1             \\ \hline\hline
\multirow{3}{*}{\makecell{Large\\Models}}& \multicolumn{1}{l|}{FoundationStereo (ViT-L)\cite{wen2025foundationstereo}} & 374.52 & 39.20 & 30.79 & 81.82 & {\ul 51.92} & {\ul 1.76} & 61.75 & \textbf{3.14} & 55.62 & {\ul 99.68} & {\ul 98.98} \\
                       & \multicolumn{1}{l|}{DEFOM (ViT-L) \cite{defom_stereo}} & 382.62 & 47.30 & {\ul 10.61} & {\ul 78.71} & 52.30 & \textbf{1.74} & {\ul 55.57} & {\ul 3.29} & {\ul 18.83} & 99.83 & 99.64 \\\cline{2-13}
                       & \multicolumn{1}{l|}{Ours (ViT-L)} & 382.62 & 47.30 & \textbf{1.51} & \textbf{31.32} & \textbf{4.24} & 1.84 & \textbf{55.05} & 3.88 & \textbf{1.41} & \textbf{6.71} & \textbf{5.66} \\ \hline\hline
\multirow{5}{*}{\makecell{Small\\Models}}& \multicolumn{1}{l|}{IGEV++ \cite{xu2025igev++}} & 14.53 & -- & 23.22 & 81.65 & 52.42 & {\ul 1.90} & 62.69 & {\ul 4.14} & 43.63 & 99.75 & 99.24 \\
                       & \multicolumn{1}{l|}{FoundationStereo (ViT-S)\cite{wen2025foundationstereo}} & 62.34 & 37.55 & 29.98 & 84.64 & 52.59 & 1.97 & 68.39 & 4.83 & 56.42 & 99.67 & 98.93 \\
                       & \multicolumn{1}{l|}{UnderwaterStereo \cite{ye2023underwater}} & 2.97 & -- & 25.17 & 81.81 & 55.51 & 7.73 & 62.61 & 8.75 & 41.31 & {\ul 99.59} & 98.67 \\
                       & \multicolumn{1}{l|}{DEFOM (ViT-S)\cite{defom_stereo} } & 43.29 & 18.51 & {\ul 4.03} & {\ul 76.10} & {\ul 49.05} & 2.31 & \textbf{49.06} & 5.52 & {\ul 6.15} & 99.84 & {\ul 91.44} \\\cline{2-13}
                       & \multicolumn{1}{l|}{Ours (ViT-S)} & 43.29 & 18.51 & \textbf{1.68} & \textbf{36.39} & \textbf{5.37} & \textbf{1.79} & {\ul 59.07} & \textbf{3.46} & \textbf{2.05} & \textbf{17.07} & \textbf{9.58} \\\hline
\end{tabular}
\label{tab:stereo_quantitative_table}
\vspace{0.2cm}
\end{table*}


\subsubsection{Model Architecture}
We finetune three different models in our stereo training: FoundationStereo, DEFOM with VIT-L weights, and DEFOM with VIT-S weights.
We elect to finetune architectures with a transformer backbone, as it has been widely shown that with large-scale datasets with heavy augmentation, vision transformer models (ViTs) scale better compared to alternative architectures~\cite{dosovitskiy2020vit,chen2021outperform}.
The selected models include both ViT-L and ViT-S weights, with the ViT-L weights targeted towards more accurate, offline tasks and the ViT-S weights targeted for real-time stereo depth estimation, which we show can be used for real-time SLAM.
The parameters used to finetune each of the three models are shown in \Table\ref{tab:configurations}.

\subsection{Baselines}
We evaluate the stereo disparity estimation and SLAM accuracy separately against different sets of baselines.

\subsubsection{Stereo Baselines}
We evaluate the proposed method against state-of-the-art stereo depth estimation networks.
These include FoundationStereo \cite{wen2025foundationstereo}, DEFOM-Stereo \cite{defom_stereo}, IGEV++ Stereo \cite{xu2025igev++} and RAFT-Stereo \cite{lipson2021raft}.
Additionally, we compare our performance with a stereo depth estimation method developed for the underwater domain, trained on data augmented with water column effects only \cite{ye2023underwater}.
For models with multiple weights, we report the best observed performance.

\subsubsection{SLAM Baselines}
For our SLAM evaluation, we compare against existing state-of-the-art methods in VIO and SLAM, including SVIn2 \cite{svin2}, a method developed for underwater state estimation, and ORB-SLAM3 \cite{orbslam3}.
For SVIn2, we compare the maps produced by the method coupled with~\cite{wang_real-time_2023}.
We additionally compare our tracking and mapping against Droid-SLAM~\cite{teed2021droid}, MASt3R-SLAM~\cite{murai2025mast3r}, and VGGT-SLAM~\cite{maggio2025vggt}, three representative methods for learning-based visual SLAM.
Finally, SLAM results are compared against TURTLMap \cite{song2024turtlmap} as a baseline acoustic-inertial fusion method.

\subsection{Evaluation Metrics}
\subsubsection{Disparity Metrics}\label{subsec:disp_metrics}

For disparity evaluation, we use the metrics from FoundationStereo~\cite{wen2025foundationstereo}. 
We report the end-point error (EPE), defined as the average per-pixel disparity error, and BP-X, the percentage of pixels whose predicted disparity deviates from the ground truth by more than X pixels. 
In addition, we report the D1 metric, which measures the percentage of pixels with disparity errors exceeding both 3 pixels and 5\% of the ground-truth disparity.

Stereo metrics are decomposed into three regions using the manually annotated foreground/background masks.
\textit{Combined} considers all pixels, \textit{On Geometry} considers pixels with valid ground-truth disparity from the photogrammetry pipeline, and \textit{Water Column} considers the manually labeled water-column pixels where the correct disparity is zero.
In all configurations, pixels labeled as foreground that lack a corresponding ground-truth depth are considered holes in the ground truth. As a result, these pixels are excluded from all metrics.

\subsubsection{Tracking Metrics}
The absolute pose error (APE) is measured between the reference and estimated trajectories.
We use the popular Evo package~\cite{grupp2017evo} to compute the tracking metrics.
Since MASt3R-SLAM and VGGT-SLAM are monocular methods, we compute trajectory metrics for these methods up to scale.

\begin{figure*}[h!tb]
    \centering
    \includegraphics[width=\linewidth]{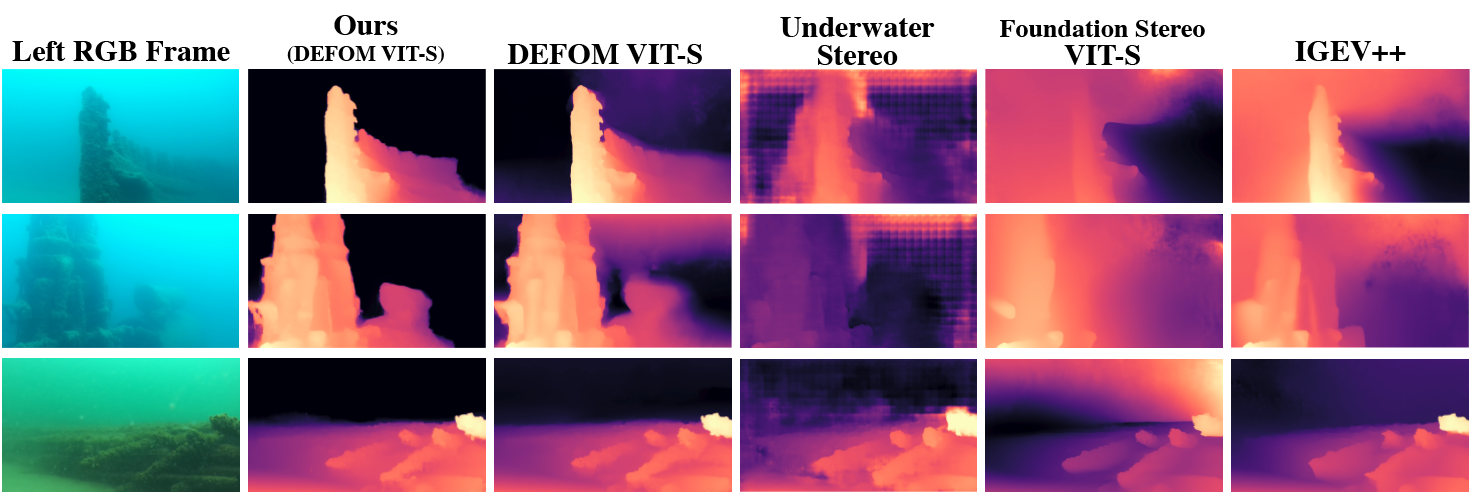}
    \caption{Qualitative comparison of the small models, including those based on the Small Vision Transformer (ViT-S) or without any transformer backbone. Ours is a fine-tuned DEFOM ViT-S, whereas Underwater Stereo and IGEV++ are both convolutional architectures. Of the baseline methods, Underwater Stereo is the only one to be trained on (simulated) underwater images.}
    \label{fig:vits_results}
\end{figure*}

\subsubsection{Mapping Metrics}
We follow the procedure from \cite{isaacson2024lonerlidarneuralrepresentations} to evaluate maps.
Each map is first downsampled to a 5cm resolution.
Then, the accuracy, completion, precision, and recall are computed.
Accuracy measures the mean distance from each point in the estimate to each point in the ground truth: a high accuracy value indicates a large number of points that are far from the ground truth.
Completion is the complementary measure that evaluates the mean distance from each ground truth point to each point in the estimate: a high completion value indicates missing portions of the reconstruction.
Precision computes the proportion of points in the estimate that are within 0.1m of a point in the ground truth.
Finally, recall computes the proportion of points in the ground truth where there is a point in the estimate within 0.1m.

\subsection{Computation Details}
Different compute systems were used for stereo training, stereo inference, metric ground truth reconstructions, and SLAM experiments.

The compute- and memory-intensive ground-truth optimizations were performed on a server equipped with dual AMD Epyc 7742 64-core CPUs and 4 TB of RAM.

The stereo networks were fine-tuned on a system with an AMD Threadripper Pro 7975WX 32-core CPU and four NVIDIA Blackwell Pro 6000 GPUs. 
Inference was performed on a separate machine with a single NVIDIA Blackwell Pro 6000.

All SLAM baselines were run on a system with an AMD Ryzen 5950X CPU and NVIDIA RTX A6000 GPU.
Our SLAM method was evaluated both on that desktop machine and on an NVIDIA Jetson Thor platform, which is suitable for deployment in the field. 
The Jetson Thor experiments were conducted on a benchtop using a stream of sensor data, and is intended to evaluate the computational feasibility of running the method on hardware suitable for onboard robotic deployment.

\renewcommand{\yes}{\checkmark}

\begin{table}[!t]
    \centering
    \footnotesize
    \caption{We ablate four features: H, haze and water-column simulation; FA, full augmentation (lighting, caustics, particles); WL, inclusion of a self-supervised warping loss; and IA, in-air data to limit drift from the initialization. Performance is measured by EPE (pixels; lower is better). All results are obtained by fine-tuning DEFOM ViT-L.}
    \begin{tabular}{cccc|ccc}
        \hline
        \multicolumn{4}{c|}{Ablation Setting} &\multirow{2}{*}{Combined} & On & Water \\[-0.5em]
        H & FA & WL & IA &  & Geom. & Column \\
        \hline
        \yes & \ding{53} & \ding{53} & \ding{53} & 6.14 & \underline{1.86} & 10.30 \\
        \yes & \yes & \ding{53} & \ding{53} & 1.68 & 1.90 & 1.79 \\
        \yes & \yes & \yes & \ding{53} & 1.57 & \underline{1.86} & 1.54 \\
        \yes & \yes & \ding{53} & \yes & \underline{1.53} & 1.91 & \underline{1.42} \\\hline
        \yes & \yes & \yes & \yes & \textbf{1.51} & \textbf{1.84} & \textbf{1.41} \\
        \hline
    \end{tabular}
    \label{tab:stereo-ablation}
    \vspace{0.2cm}
\end{table}

\section{Stereo Evaluation Results}\label{sec:results}

\Table \ref{tab:stereo_quantitative_table} provides a quantitative comparison of our trained models against baseline methods for stereo depth estimation, where we report the best-performing models from the three that we fine-tuned.
We compare our models against three different model classifications: models with a ViT-L backbone~\cite{wen2025foundationstereo,defom_stereo}, ViT-S backbone~\cite{wen2025foundationstereo,defom_stereo}, and models that lack a ViT component~\cite{xu2025igev++,ye2023underwater}.
Note that our method uses a DEFOM-Stereo architecture (large and small models) with our proposed train-time augmentations and loss function, as it was the best-performing model from our fine-tuning.

In quantitative evaluations, the most drastic improvement seen in our proposed model is the ability to correctly identify the water column in the depth prediction, as noted in the \textit{Water Column Only} evaluation.
Our models perform competitively with or better than existing methods in the on-geometry evaluation. The \textit{Combined} metric, which encapsulates the overall performance of each model, demonstrates that our proposed training improves stereo depth estimation in underwater environments.

\Figure\ref{fig:vits_results} shows qualitative comparisons of the small models and \Figure \ref{fig:vitl_results} shows qualitative comparisons of the large models. 
The qualitative results demonstrate that existing stereo depth estimation methods fail to transfer to underwater settings, producing either noisy depth estimates or systematic errors such as failure to estimate depth in the water column.
Our model is able to remove the water column from challenging images across different water column types and haze levels.

\begin{figure}[h!tp]
    \vspace{-0.25cm}

    \centering
    \includegraphics[width=1.0\linewidth]{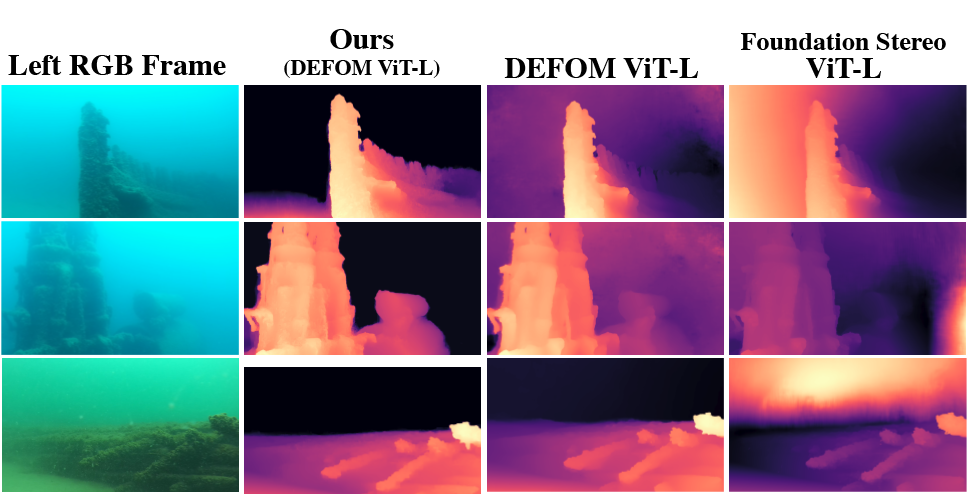}
    \caption{Qualitative results of the Large Vision Transformer (ViT-L) models. Ours, based on DEFOM ViT-L, balances strong on-geometry performance with effective background removal.}
    \label{fig:vitl_results}
    \vspace{-0.25cm}
\end{figure}

\begin{table}[t!]
\centering
\footnotesize
\caption{Estimated trajectories are compared against the ground truth. The reported metric is Root Mean Square Absolute Pose Error (APE). Each result is the median of five runs. MASt3R-SLAM is a monocular method, so scale-invariant trajectory evaluations were conducted for MASt3R-SLAM only. Lower is better.}
\begin{tabular}{l||c|c|c|c}
\hline
Method & Platform & \multicolumn{1}{c|}{Engine} & \multicolumn{1}{c|}{Boiler} & \multicolumn{1}{c}{Long} \\
\hline
DROID-SLAM \cite{teed2021droid} & CPU+GPU & \textbf{0.164} & 3.121 & 5.794 \\
MASt3R SLAM \cite{murai2025mast3r} & CPU+GPU & 0.395 & \underline{0.462} & 1.963 \\
ORB-SLAM3 \cite{orbslam3} & CPU & 1.662 & 2.844 & 6.783 \\
SVIn2 \cite{svin2} & CPU & 1.085 & 2.413 & 2.774 \\
TURTLMap \cite{song2024turtlmap} & CPU & 0.399 & 0.528 & \underline{1.312} \\
\hline
Ours & CPU+GPU & \underline{0.216} & \textbf{0.246} & \textbf{0.444} \\\hline
\end{tabular}
\vspace{0.2cm}
\label{tab:slam_results}
\end{table}

\subsection{Ablation Study on Stereo Depth Estimation}
We perform an ablation over the key components of the proposed stereo network fine-tuning.
The evaluation of the ablation is done by comparing the EPE on the \textit{On Geometry}, \textit{Water Column}, and \textit{Combined} metrics.
The ablation results are shown in \Table \ref{tab:stereo-ablation}.

Each ablation removes a single component while keeping all others fixed, allowing isolation of its effect on geometry estimation and water-column estimation.
The ablation reveals three distinct effects.
Full augmentations (FA) are necessary to reduce predictions in the water column. 
The self-supervised warping loss (WL) primarily improves depth accuracy on observed geometry, while the inclusion of in-air data (IA) prevents degradation under heavy augmentation. 
The combination of WL and IA yields the best trade-off between geometric fidelity and incorrect predictions on the water column.

\section{SLAM Evaluation Results}
We evaluate the trajectories from each SLAM method.
ORB-SLAM3 is excluded from mapping metrics since the sparse keypoint map is not intended to be used as a structural map.
We note that VGGT-SLAM, which is a calibration-free monocular SLAM method \cite{maggio2025vggt}, was unable to produce a trajectory estimate on any of the sequences and is thus excluded from the result tables.
For SVIn2, the mapping metrics are obtained using the dense fusion method proposed in the authors' follow-up work \cite{wang_real-time_2023}.

\subsection{Trajectory Tracking Evaluation}
Results of the SLAM trajectory evaluation are shown in Table \ref{tab:slam_results}.
Our proposed method shows the best performance on the \textit{Boiler} and \textit{Long} SLAM sequences, and second best on the \textit{Engine} sequence.
TURTLMap, which does not apply global registration, produces reasonable trajectories on all sequences. Yet, without a mechanism to mitigate accumulated drift, it underperforms against methods that are able to successfully register against a global map.
Of the purely vision-based methods, DROID-SLAM and MASt3R-SLAM perform the best.
This indicates that even without fine-tuning, learning-based SLAM is better equipped to deal with the challenges of the underwater domain compared to ORB-SLAM3, which uses carefully tuned feature-matching heuristics.
SVIn2 converges on all sequences but drifts significantly. We hypothesize that this drift results from the IMU initialization procedure in SVIn2, which benefits from the robot starting at rest. In our field trials, it was not feasible for the robot to be stationary due to dynamic environmental conditions.

We perform an ablation study of the sensors used for tracking, shown in \Table\ref{tab:tracking_ablation_results}.
The results demonstrate that the DVL plays a critical role in tracking, as disabling it leads to substantial drift across all sequences.
Disabling the camera corresponds to running TURTLMap~\cite{song2024turtlmap} and degrades performance, particularly on the \textit{Long} sequence, demonstrating the benefit of global registration for reducing accumulated drift.
In contrast, the barometer has a relatively small effect on tracking performance: disabling it produces similar results on \textit{Engine} and \textit{Boiler} and slightly improves performance on \textit{Long}.
We hypothesize that this is due to noise or bias in the barometer measurements, which can occasionally introduce inconsistent depth constraints when the remaining sensors already provide sufficient information to estimate the trajectory.

\begin{figure*}[h!t!]
    \centering
    \includegraphics[width=1.0\linewidth]{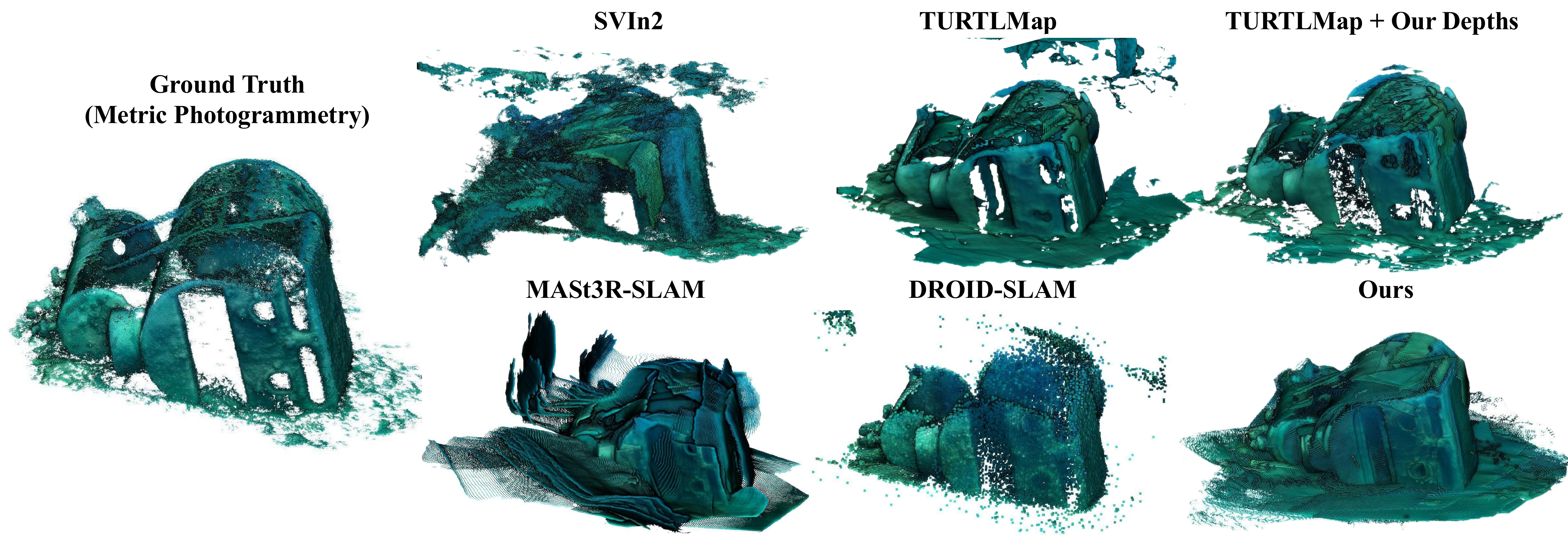}
    \caption{Qualitative comparison of the maps generated by each method on the \textit{Boiler} sequence. Methods that produce vastly incorrect maps due to their incorrect pose estimates (specifically, DroidSLAM~\cite{teed2021droid}, SVIn2~\cite{svin2,wang_real-time_2023}, and TURTLMap~\cite{song2024turtlmap}) are cropped for visualization purposes. The SVIn2 image refers to running the SVIn2 \cite{svin2} for trajectory estimation and using the dense mapping method from \cite{wang_real-time_2023}. Our map produces a more accurate, cleaner, and more complete map than those produced by the baseline methods.}
    \label{fig:maps}
\end{figure*}
\newcommand{\xmark}{\ding{53}}
\newcommand{\cmark}{\yes}

\begin{table}[t!]
\caption{Ablation results for disabling individual sensors from the SLAM pipeline. Disabling the camera corresponds to running TURTLMap \cite{song2024turtlmap}. The reported metric is RMS APE; each result is the median of five runs.}
\centering
\footnotesize
\begin{tabular}{ccc||ccc}
\toprule
Camera & DVL & Baro. & Engine & Boiler & Long \\
\midrule
\cmark & \cmark & \xmark & \underline{0.226} & \textbf{0.246} & \textbf{0.386} \\
\cmark & \xmark & \cmark & 14.439 & 102.070 & 450.411 \\
\xmark & \cmark & \cmark & 0.399 & 0.528 & 1.312 \\
\midrule
\cmark & \cmark & \cmark & \textbf{0.216} & \textbf{0.246} & \underline{0.444} \\
\bottomrule
\end{tabular}
\label{tab:tracking_ablation_results}
\end{table}

\begin{table}[t]
\centering
\caption{Maps were evaluated using photogrammetry ground-truth. The TURTL Base column indicates accumulating depths from the
TURTLMap trajectory estimates using a baseline DEFOM ViT-S
stereo depth estimator. In contrast, TURTL Ours indicates using the TURTLMap trajectories and our proposed DEFOM ViT-S stereo depth estimator. All numbers are the median of five runs. For accuracy and completeness, lower is better. For precision and recall, higher is better.}
\setlength{\tabcolsep}{3pt}
\resizebox{\columnwidth}{!}{%
\scriptsize
\begin{tabular}{c l | c c c c | c c}
 &  & DROID- & MASt3R- & \multirow{2}{*}{SVIn2} & TURTL & TURTL & \multirow{2}{*}{Ours} \\[-0.2em]
 &  & SLAM & SLAM &  & Base & Ours &  \\
\hline
\multirow{4}{*}{\rotatebox{90}{Boiler}}
 & Acc.  & \textbf{0.29} & 0.44 & 1.75 & 1.05 & 0.48 & \underline{0.30} \\
 & Comp.  & 0.39 & \underline{0.09} & 0.16 & 0.16 & 0.15 & \textbf{0.04} \\
 & Prec.  & 0.17 & \underline{0.32} & 0.14 & 0.25 & 0.26 & \textbf{0.39} \\
 & Rec.  & 0.07 & \underline{0.82} & 0.63 & 0.42 & 0.41 & \textbf{0.96} \\
\hline
\multirow{4}{*}{\rotatebox{90}{Engine}}
 & Acc.  & \textbf{0.04} & 0.53 & 0.67 & \underline{0.09} & 0.14 & 0.34 \\
 & Comp.  & 0.35 & 0.20 & \underline{0.17} & 0.28 & 0.60 & \textbf{0.05} \\
 & Prec.  & \textbf{0.98} & 0.29 & 0.13 & \underline{0.74} & 0.55 & 0.52 \\
 & Rec.  & 0.14 & \underline{0.79} & 0.66 & 0.46 & 0.32 & \textbf{0.98} \\
\hline
\multirow{4}{*}{\rotatebox{90}{Long}}
 & Acc.  & 1.63 & 0.82 & 1.86 & \textbf{0.19} & \textbf{0.19} & 0.50 \\
 & Comp.  & 2.08 & 5.50 & \underline{0.18} & 0.25 & 0.25 & \textbf{0.03} \\
 & Prec.  & 0.01 & 0.15 & 0.09 & \textbf{0.58} & \underline{0.48} & 0.47 \\
 & Rec.  & 0.00 & 0.26 & \underline{0.62} & 0.39 & 0.38 & \textbf{0.99} \\
\hline
\end{tabular}%
}
\vspace{0.2cm}
\label{tab:mapping}
\end{table}

\subsection{Mapping Evaluation}
Quantitative map evaluations are presented in Table \ref{tab:mapping}.
We present the mapping metrics from the trajectories evaluated on the desktop platform, and the results indicate that SurfSLAM has comparable or better performance across these metrics.
On the completion and recall metrics, \algname demonstrates significant improvements, indicating more complete reconstructions.
For the accuracy and precision metrics, performance is degraded on the \textit{Engine} and \textit{Long} sequences due to a few erroneous pose estimates, resulting in a misaligned projected depth map.
This is supported by the qualitative results shown in \Figure\ref{fig:maps}, where the map produced by \algname shows consistently more accurate measurements and fewer measurements misaligned with the larger reconstructed map.
On the \textit{Long} sequence, which features the most low-texture regions, the advantages of \algname are most apparent, producing the most complete reconstruction.
DROID-SLAM excels in accuracy and precision on the feature-rich \textit{Engine} scene. 
Yet, the completeness and recall are comparatively worse than those of \algname, indicating less complete maps.
SVIn2's dense fusion struggles, likely due to its difficulty with trajectory estimation (See Table~\ref{tab:slam_results}). 
Running TURTLMap with our stereo network is comparable to \algname, with a consistent split across metrics.
We show consistent improvement in the completeness of the reconstructions with comparable accuracy on longer sequences.
Global registration trades a small amount of local accuracy for substantially more complete reconstructions.

Qualitative mapping results are shown in \Figure\ref{fig:maps}.
Our method produces the most complete and geometrically consistent reconstruction among the compared approaches.
For DroidSLAM, the trajectory estimate diverges as the robot traverses the boiler and loses tracking, leading to an incomplete map, consistent with the quantitative results.
For MASt3R-SLAM, despite achieving the second-best trajectory accuracy, depth prediction errors introduce noticeable artifacts and reduce map accuracy.
For SVIn2 and TURTLMap, accumulated drift distorts and misaligns the reconstruction: SVIn2 exhibits pronounced distortion on the left side of the map, while TURTLMap shows missing geometry due to depth misalignment, causing valid observations to be filtered during mapping.
Finally, the map produced by TURTLMap using our proposed depths is comparable to the maps produced by TURTLMap alone.

\subsection{Compute Comparison and Runtime}\label{sec:runtime}
\begin{table}[t!]
\centering
\footnotesize
\caption{We compare the tracking performance of our method on each of the three sequences on two platforms, a high-performance desktop machine and an NVIDIA Jetson Thor. The reported metric is Root Mean Square Absolute Pose Error (APE).}
\begin{tabular}{l||c|c|c}
\hline
Platform & \multicolumn{1}{c|}{Engine} & \multicolumn{1}{c|}{Boiler} & \multicolumn{1}{c}{Long} \\
\hline
Desktop & 0.216 & 0.246 & 0.444 \\
Jetson Thor & 0.171 & 0.216 & 0.333 \\\hline
\end{tabular}
\vspace{0.2cm}
\label{tab:tracking-compute}
\end{table}

This section compares the performance of our method between the desktop platform and the NVIDIA Jetson Thor.

Table \ref{tab:tracking-compute} shows that \algname achieves comparable performance on both desktop and embedded platforms, reinforcing its suitability for field deployment. On  all three sequences, the Jetson yields slightly lower tracking error than the desktop. Because the Jetson drops more frames during stereo matching (see Table \ref{tab:pipeline_stats}), a plausible explanation is that it skips a frame that would otherwise induce an erroneous registration on the desktop platform.

We present the runtime of key components of the algorithm in Table \ref{tab:pipeline_stats}.
On each of the platforms used to evaluate the system, we report two key timing metrics: time for stereo inference and time to solve the backend optimization.
In practice, several of the components run in parallel.
To ensure real-time operation, each component drops stale data.
Table \ref{tab:pipeline_stats} also reports statistics on how many frames are dropped within the system.
On the desktop platform, 98.4\% of input stereo frames are processed by the stereo inference module.
All of those frames are processed by the registration module to search for global registrations.
On the Jetson platform, the same 98.4\% of the input stereo frames are processed. 
However, due to the slower stereo inference time, 22.8\% of frames processed by the stereo matcher become stale and are dropped before registration is attempted. 

\section{Discussion \& Conclusion}\label{sec:conclusion}
This paper presents a comprehensive methodology for stereo-based state estimation and scene reconstruction in challenging underwater environments.
We first introduced a sim-to-real training pipeline for underwater stereo disparity estimation and demonstrated that it outperforms the state-of-the-art in underwater domains.
Next, we presented \algname, a method that fuses stereo images with IMU, DVL, and barometer measurements to perform long-term state estimation and mapping in real-world underwater environments.
In addition to the algorithm contributions of this paper, we present two new datasets for underwater stereo perception and underwater SLAM. The first dataset comprises both a curated dataset of scenes that mimic a variety of underwater environments and an augmentation pipeline that randomizes underwater imaging effects. We also present a comprehensive dataset of real-world shipwrecks.

The results of our experiments demonstrate state-of-the-art performance on stereo estimation in the underwater domain.
While several related works simulate underwater haze, we additionally simulate directional lighting, caustics, and suspended particles.
Experiments demonstrate that this comprehensive simulation improves performance over the haze-only simulation.
Further, while baseline methods most often train either on simulated data or via self-supervised training on real underwater images, we demonstrate that a combined approach yields superior results.

The results of our SLAM evaluation support the approach of relying on acoustic and inertial sensors as the primary local tracker, while using a highly accurate but slow vision model to correct drift when possible. 
The trajectory evaluation results support the conclusion that this leads to stronger performance in challenging environments than methods that rely on feature tracking, which is susceptible to failure in low-texture underwater scenes.

A limitation of our existing evaluations arises from the difficulty of reconstructing metric ground truth from images under highly turbid conditions.
While we hypothesize that our sim-to-real pipeline will have a greater impact in more turbid water, it is challenging to test this hypothesis because photogrammetry methods cannot reconstruct ground truth in such conditions.
A potential avenue for future work is to use other sensors, such as 3D SONAR or Underwater LiDAR, to generate high-quality ground truth in such environments.

An avenue for future work to improve \algname is to augment the SLAM pipeline with bundle adjustment rather than relying on frame-to-frame registration.
In practice, this is challenging in our target domain, which includes large, textureless regions that can push the stability of existing bundle-adjustment methods. 
\begin{table}[t!]
\centering
\footnotesize
\caption{We report the runtime of both stereo inference and the backend optimization (mean +/- standard deviation), averaged across all sequences and 5 trials. During real-time operation of the system, stale frames are dropped to prevent the system from falling behind. For instance, an N\% registration drop rate means that of all the frames the registration module received, it discarded N\% of the frames and processed the rest.}
\label{tab:pipeline_stats}
\begin{tabular}{l|l|rr}
\hline
Group & Metric & Ours (Desktop) & Ours (Jetson) \\
\hline
\multirow{2}{*}{Timing} & Stereo Inference & $359.3 \pm 26.4$ ms & $786.1 \pm 51.2$ ms\\
 & Optimization & $6.7 \pm 3.4$ ms & $9.9 \pm 5.1$ ms \\
\hline
\multirow{3}{*}{Drops} & Stereo & $1.6$ \% & $1.6$ \% \\
 & Registration & $0.0$ \% & $21.5$ \% \\
 \cline{2-4}
 & Overall & $1.6$ \% & $22.8$ \% \\
\hline
\end{tabular}
\end{table}

{
    \small
    \bibliographystyle{ieeenat_fullname}
    \bibliography{main}
}
\end{document}